\documentclass[10pt,journal,compsoc]{IEEEtran}
\ifCLASSOPTIONcompsoc
\else
\fi
\ifCLASSINFOpdf
\else
\fi

\usepackage[bookmarks=false,colorlinks=true,urlcolor=black,citecolor={green!60!black}]{hyperref}   
\usepackage[english]{babel}
\usepackage[square,numbers]{natbib}
\usepackage{wrapfig}
\usepackage{subcaption}
\usepackage{bbding}
\usepackage{utfsym}
\usepackage{tabularx}
\usepackage{booktabs}% http://ctan.org/pkg/booktabs
\usepackage{multirow}
\usepackage{pifont}
\usepackage[thinlines]{easytable}
\usepackage{enumitem}
\usepackage{url}
\usepackage{hyperref}
\usepackage{threeparttable}
\usepackage{multirow}
\usepackage{graphicx}
\usepackage{adjustbox}
\usepackage[labelformat=simple]{subcaption}
\usepackage{caption}
\usepackage{lineno}
\usepackage{xspace}
\usepackage{enumitem}
\usepackage[most]{tcolorbox}
\newtheorem{definition}{Definition}[section]
\newtheorem{theorem}{Theorem}[section]

\newtheorem{assumption}[theorem]{Assumption}

\makeatletter
\DeclareRobustCommand\onedot{\futurelet\@let@token\bmv@onedotaux}
\def\bmv@onedotaux{\ifx\@let@token.\else.\null\fi\xspace}

\def\eg{\emph{e.g}\onedot} \def\Eg{\emph{E.g}\onedot}
\def\ie{\emph{i.e}\onedot} \def\Ie{\emph{I.e}\onedot}
\def\cf{\emph{c.f}\onedot} \def\Cf{\emph{C.f}\onedot}
\def\etc{\emph{etc}\onedot} \def\vs{\emph{vs}\onedot}
\def\wrt{w.r.t\onedot}
\def\dof{d.o.f\onedot}
\def\aka{a.k.a\onedot }
\def\etal{\emph{et al}\onedot}
\def\bigoh{\mathcal{O}}

\ifdefined \GramaCheck
  \newcommand{\CheckRmv}[1]{}
  \newcommand{\figref}[1]{Figure 1}%
  \newcommand{\tabref}[1]{Table 1}%
  \newcommand{\secref}[1]{Section 1}
  \renewcommand{\eqref}[1]{Equation 1}
\else
  \newcommand{\CheckRmv}[1]{#1}
  \newcommand{\figref}[1]{Fig.~\ref{#1}}%
  \newcommand{\tabref}[1]{Table~\ref{#1}}%
  \newcommand{\secref}[1]{Section~\ref{#1}}
  \renewcommand{\eqref}[1]{Equation~(\ref{#1})}
\fi

\begin{document}
%
% paper title
% Titles are generally capitalized except for words such as a, an, and, as,
% at, but, by, for, in, nor, of, on, or, the, to and up, which are usually
% not capitalized unless they are the first or last word of the title.
% Linebreaks \\ can be used within to get better formatting as desired.
% Do not put math or special symbols in the title.
\title{A Survey of Trustworthy Federated Learning with Perspectives on Security, Robustness, and Privacy}
%
%
% author names and IEEE memberships
% note positions of commas and nonbreaking spaces ( ~ ) LaTeX will not break
% a structure at a ~ so this keeps an author's name from being broken across
% two lines.
% use \thanks{} to gain access to the first footnote area
% a separate \thanks must be used for each paragraph as LaTeX2e's \thanks
% was not built to handle multiple paragraphs
%
%
%\IEEEcompsocitemizethanks is a special \thanks that produces the bulleted
% lists the Computer Society journals use for "first footnote" author
% affiliations. Use \IEEEcompsocthanksitem which works much like \item
% for each affiliation group. When not in compsoc mode,
% \IEEEcompsocitemizethanks becomes like \thanks and
% \IEEEcompsocthanksitem becomes a line break with idention. This
% facilitates dual compilation, although admittedly the differences in the
% desired content of \author between the different types of papers makes a
% one-size-fits-all approach a daunting prospect. For instance, compsoc 
% journal papers have the author affiliations above the "Manuscript
% received ..."  text while in non-compsoc journals this is reversed. Sigh.

\author{Yifei Zhang$^*$,
        Dun Zeng$^*$,
        Jinglong Luo$^*$,
        Zenglin Xu$^\dagger$
        Irwin King$^\dagger$% <-this % stops a space
\IEEEcompsocitemizethanks{
\IEEEcompsocthanksitem Yifei Zhang and Irwin King was with the Department of Computer Science and Engineering, The Chinese University of Hong Kong, ShaTin, N.T., Hong Kong SAR, China. Email: \{yfzhang,king\}@cse.cuhk.hk
\IEEEcompsocthanksitem Zeng Dun was with the University of Electronic Science and Technology of China and Peng Cheng Lab. Email: zengdun@std.uestc.edu.cn
\IEEEcompsocthanksitem Jinglong Luo  and Zenglin Xu was Harbin Institute of Technology, Shenzhen and Peng Cheng Lab Email: luojl@pcl.ac.cn and xuzenglin@hit.edu.cn
% \IEEEcompsocthanksitem Yifei Zhang Dun Zeng and Jinglong Luo contribute equally. Corresponding Authors are Zengling Xu and Irwin King
}
\thanks{Yifei Zhang Dun Zeng and Jinglong Luo contribute equally. Corresponding Authors are Zengling Xu and Irwin King}}

\IEEEtitleabstractindextext{%
\begin{abstract}
Trustworthy artificial intelligence (AI) technology has revolutionized daily life and greatly benefited human society. Among various AI technologies, Federated Learning (FL) stands out as a promising solution for diverse real-world scenarios, ranging from risk evaluation systems in finance to cutting-edge technologies like drug discovery in life sciences. However, challenges around data isolation and privacy threaten the trustworthiness of FL systems. Adversarial attacks against data privacy, learning algorithm stability, and system confidentiality are particularly concerning in the context of distributed training in federated learning. Therefore, it is crucial to develop FL in a trustworthy manner, with a focus on security, robustness, and privacy. In this survey, we propose a comprehensive roadmap for developing trustworthy FL systems and summarize existing efforts from three key aspects: security, robustness, and privacy. We outline the threats that pose vulnerabilities to trustworthy federated learning across different stages of development, including data processing, model training, and deployment. To guide the selection of the most appropriate defense methods, we discuss specific technical solutions for realizing each aspect of Trustworthy FL (TFL). Our approach differs from previous work that primarily discusses TFL from a legal perspective or presents FL from a high-level, non-technical viewpoint.

\end{abstract}

% Note that keywords are not normally used for peerreview papers.
}

% make the title area
\maketitle

\DeclareRobustCommand\onedot{\futurelet\@let@token\bmv@onedotaux}
\def\bmv@onedotaux{\ifx\@let@token.\else.\null\fi\xspace}
\def\eg{\emph{e.g}\onedot} \def\Eg{\emph{E.g}\onedot}
\def\ie{\emph{i.e}\onedot} \def\Ie{\emph{I.e}\onedot}
\def\cf{\emph{c.f}\onedot} \def\Cf{\emph{C.f}\onedot}
\def\etc{\emph{etc}\onedot} \def\vs{\emph{vs}\onedot}
\def\wrt{w.r.t\onedot}
\def\dof{d.o.f\onedot}
\def\aka{a.k.a\onedot }
\def\etal{\emph{et al}\onedot}
\def\bigoh{\mathcal{O}}

\section{Introduction~\label{sec:intro}}
\subsection{From Trustworthy AI to Trustworthy Federated Learning}
Trustworthy AI has recently received increased attention due to the need to avoid the adverse effects that AI could have on people~\citep{zhang2022trustworthy,truex2019demystifying175, liu2021trustworthy, song2022towards, rizoiu2018debatenight}. There have been ongoing efforts to promote trustworthy AI, so that people can fully trust and live in harmony with AI technologies. Among various AI technologies~\citep{zhang2022costa,song2021semi,zhang-zhu-2019-doc2hash,zhang2020discrete,zhang2022spectral,zhang2022graph, chen2021attentive,chen2021hyper, chen2022learning, bgch}, Federated Learning (FL) stands out due to the demand for data privacy and data availability in distributed environments. The core idea of FL is to generate a collaboratively trained global learning model without sharing the data owned by the distributed clients~\citep{yin2021comprehensive,rodriguez2023survey, zhang2020additively}. Since its introduction in 2016~\citep{mcmahan2017communication121}, FL has been widely used in various areas, including finance~\citep{househ2021multiple}, health~\citep{rau2021palgrave}, business, and entertainment. However, FL is vulnerable to adversarial attacks that are mainly focused on impairing the learning model or violating data privacy, posing significant threats to FL in safety-critical environments. Therefore, people increasingly expect FL to be private, reliable, and socially beneficial enough to be trusted. In this survey, we provide a summary of these efforts from the perspective of Trustworthy Federated Learning (TFL).

\begin{figure*}[t]
\begin{subfigure}[b]{0.68\textwidth}
         \centering
    \includegraphics[width=\textwidth]{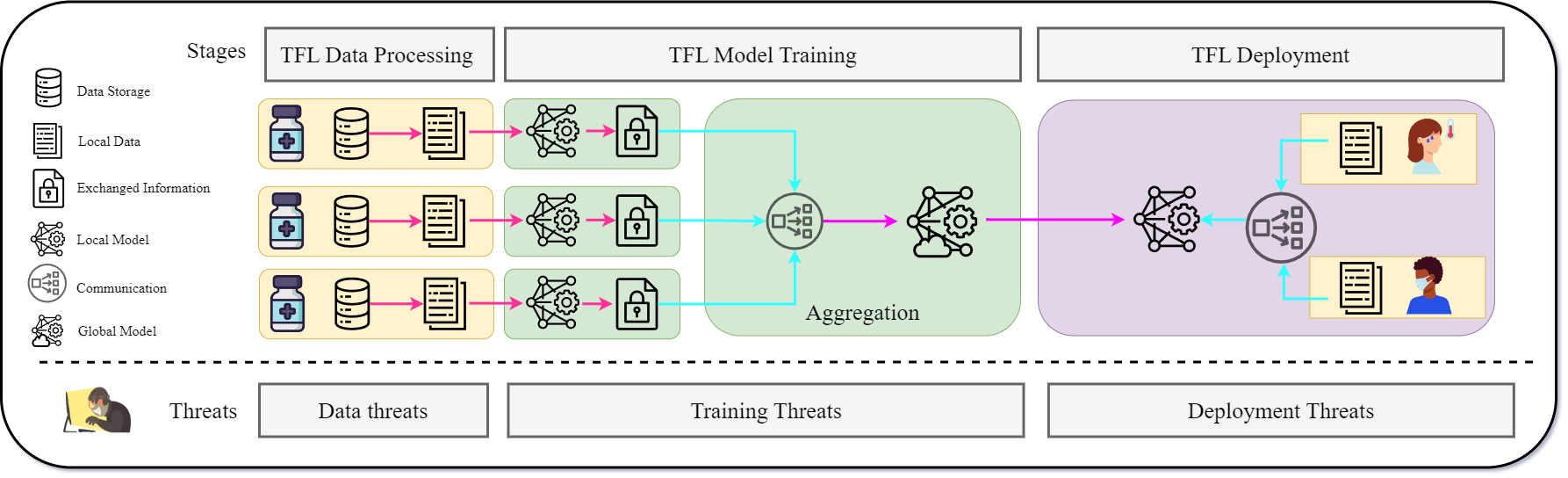}
    \caption{The development stage of TFL\label{fig:stage}.}
\end{subfigure}
\begin{subfigure}[b]{0.26\textwidth}
         \centering
    \includegraphics[width=\textwidth]{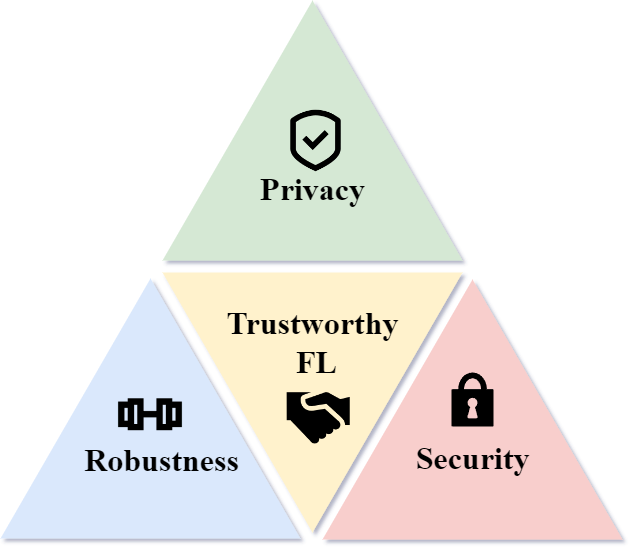}
    \caption{Aspects in TFL\label{fig:aspect}}
\end{subfigure}
\caption{The overall framework of Trustworthy Federated Learning (TFL).}
% \vspace{-0.3cm}
\end{figure*}

The trustworthiness of the system is based on the concept of "trust," which makes the system "worthy" of being relied on. Trust can be defined as "the relationship between a trustor and a trustee: the trustor trusts the trustee." In the context of TFL, we define the trustee as the FL models or systems, and the trustor as the participants in FL, users, and regulators. From the perspective of software development, an FL model is trustworthy if it is free of threats in different development stages of an FL model (\ie, data processing, model training, and deployment), with three key aspects of trustworthiness, as we will discuss later. We define TFL as being safe and private for processing personal data (Privacy), stable and robust for training the model (Robustness), and confidential and correct for deploying systems (Security), as shown in \figref{fig:stage}.

\begin{figure*}
    \centering
    \includegraphics[width=\textwidth]{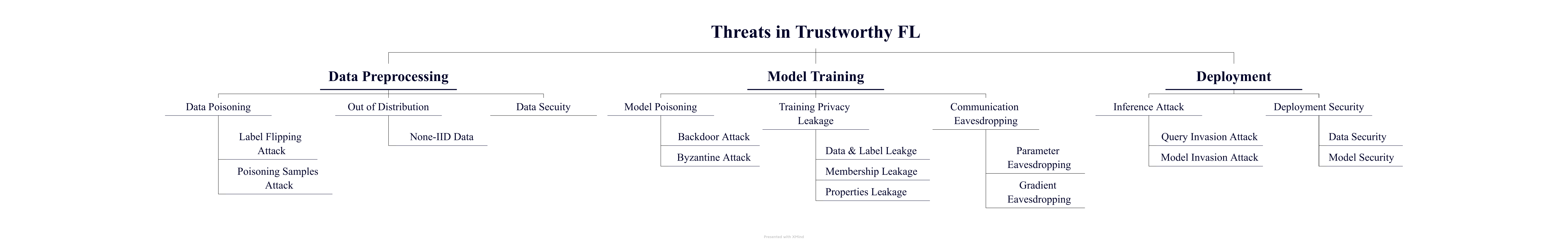}
    \caption{Threats in different stages of Trustworthy Federated Learning.\label{fig:threats}}
    % \vspace{-0.5cm}
\end{figure*}

\begin{table*}[t]
\renewcommand{\arraystretch}{3}
\caption{The proposed taxonomy of Trustworthy Federated Learning\label{tab:taxonomy}}
\resizebox{\textwidth}{!}{
\begin{tabular}{cc|cc|cc|cc}
\toprule
 &                            & \multicolumn{2}{c|}{\textbf{Data Processing}}                                            & \multicolumn{2}{c|}{\textbf{Model Traing}}                                               & \multicolumn{2}{c}{\textbf{Deployment}}                                               \\
 \cline{3-8} 
 &                            & \multicolumn{1}{c|}{Treats}                                 & Defense          & \multicolumn{1}{c|}{Treats}                                 & Defense          & \multicolumn{1}{c|}{Treats}                               & Defense          \\ 
 \cline{2-8} 
 & \multirow{2}{*}{\textbf{Security}}   & \multicolumn{1}{c|}{\multirow{5}{*}{\begin{minipage}[c]{4cm}
Information Leakage
\begin{itemize}[align=parleft,left=0pt..1em]
    \item Data \& Label Leakage~\citep{sannai2018reconstruction153,zhao2020idlg_p_223,zhu2019deep228}
    \item Membership Leakage ~\citep{dwork2010difficulties42, lu2020sharing114, mcmahan2017learning123, nasr2019comprehensive129, nasr2018machine130, pustozerova2020information139, shokri2017membership160}
    \item Properties Leakage ~\citep{ateniese2013hacking8, fredrikson2014privacy54, mcmahan2017learning123}
\end{itemize}
\end{minipage}}}          & \multirow{2}{*}{\begin{minipage}[c]{3cm}
\begin{itemize}[align=parleft,left=0pt..1em]
    \item Cryptography-based Methods
    ~\citep{mohassel2017secureml, agrawal2019quotient, patra2021aby2, kelkar2022secure, Mohassel2018ABY3, Wagh2019SecureNN, wagh2020falcon, ryffel2020ariann, patra2020blaze, li2019privpy, chaudhari2019trident,  dalskov2021fantastic,  byali2020flash,  koti2021swift, koti2021tetrad}
\end{itemize}
\end{minipage}}  & \multicolumn{1}{c|}{\multirow{2}{*}{\begin{minipage}[c]{4cm}
    Gradient Inversion Attack~\citep{GIV1,zhao2020idlg_p_223, GIV2,GIV4,zhu2019deep228}
\end{minipage}}}                      & \multirow{2}{*}{\begin{minipage}[c]{3cm}
\begin{itemize}[align=parleft,left=0pt..1em]
    \item Cryptography-based Methods
    ~\citep{bonawitz2017practical, mandal2018nike, dong2020eastfly,beguier2020safer,xu2019verifynet_p_198,guo2020v,aono2017privacy,zhang2020batchcrypt,zhang2020privacy,cheng2021separation, zhao2020smss}
\end{itemize}
\end{minipage}}
&\multicolumn{1}{c|}{\multirow{2}{*}{\begin{minipage}[c]{3cm}
Deployment Security
\begin{itemize}[align=parleft,left=0pt..1em]
    \item Data security
    \item Model security
\end{itemize}
\end{minipage}}} & \multirow{2}{*}{\begin{minipage}[c]{3cm}
\begin{itemize}[align=parleft,left=0pt..1em]
    \item Cryptography-based Methods
    ~\citep{hesamifard2017cryptodl, gilad2016cryptonets,fenner2020privacy,chen2022x,rouhani2018deepsecure,liu2017oblivious,riazi2018chameleon,juvekar2018gazelle, mishra2020delphi, riazi2019xonn,chaudhari2019astra, rathee2020cryptflow2,huang2022cheetah,zhu2022securebinn,rathee2021sirnn}
\end{itemize}
\end{minipage}} \\
 &                            & \multicolumn{1}{c|}{}                                       &                   & \multicolumn{1}{c|}{}                                       &                   & \multicolumn{1}{c|}{}                                     &                   \\ 
\cline{2-2}\cline{4-8}
 & \multirow{3}{*}{\textbf{Privacy}}   & \multicolumn{1}{c|}{}                                       & \multirow{3}{*}{\begin{minipage}[c]{3cm}

\begin{itemize}[align=parleft,left=0pt..1em]
    \item Anonymization Methods ~\citep{lu2020sharing114,truex2019demystifying175,fung2018dancing_p_56,song2020analyzing_p_165,xie2019slsgd_p_197,choudhury2020syntactic_p_32,zhao2021anonymous_p_222}
\end{itemize}
\end{minipage}} & 
\multicolumn{1}{c|}{\multirow{3}{*}{\begin{minipage}[c]{4cm}
Training Phase Privacy Leakage
\begin{itemize}[align=parleft,left=0pt..1em]
    \item Data \& Label Leakage~\citep{sannai2018reconstruction153,zhao2020idlg_p_223,zhu2019deep228}
    \item Membership Leakage ~\citep{dwork2010difficulties42, lu2020sharing114, mcmahan2017learning123, nasr2019comprehensive129, nasr2018machine130, pustozerova2020information139, shokri2017membership160}
    \item Properties Leakage ~\citep{ateniese2013hacking8, fredrikson2014privacy54, mcmahan2017learning123}
\end{itemize}
\end{minipage}}}
& \multirow{3}{*}{\begin{minipage}[c]{4cm}
\begin{itemize}[align=parleft,left=0pt..1em]
    \item DP-based methods~\citep{agarwal2018cpsgd_p_3,hao2019towards_p_70,hu2020personalized_p_78,lu2019differentially_p_116,lu2020decentralized_p_115,zhao2020local_p_225}
    \item Perturbation-based Method~\citep{hao2019efficient69,liu2020adaptive_p_110,wei2020federated_p_193}
\end{itemize}
\end{minipage}} & \multicolumn{1}{c|}{\multirow{3}{*}{\begin{minipage}[c]{3cm}
Inference attack
\begin{itemize}[align=parleft,left=0pt..1em]
    \item Query-based attack \citep{ateniese2013hacking8, fredrikson2014privacy54, mcmahan2017learning123}
    \item Model-based attack \citep{shokri2017membership160, melis2019exploiting}
\end{itemize}
\end{minipage}}}    & \multirow{3}{*}{\begin{minipage}[c]{4cm}
\begin{itemize}[align=parleft,left=0pt..1em]
    \item DP-based methods~\citep{mcmahan2017learning123,zhu2019deep228}
    \item Perturbation-based Method~\citep{wei2020framework194,chang2019cronus_p_20,feng2020practical_p_52}
\end{itemize}
\end{minipage}} \\
 &                            & \multicolumn{1}{c|}{}                                       &                   & \multicolumn{1}{c|}{}                                       &                   & \multicolumn{1}{c|}{}                                     &                   \\
 &                            & \multicolumn{1}{c|}{}                                       &                   & \multicolumn{1}{c|}{}                                       &                   & \multicolumn{1}{c|}{}                                     &                   \\ 
 \cline{2-8} 

 & \multirow{3}{*}{\textbf{Robustness}} & \multicolumn{1}{c|}{Non-IID Data~\citep{zhao2018federated,hsieh2020non}}              & \multirow{3}{*}{} & \multicolumn{1}{c|}{\multirow{3}{*}{\begin{minipage}[c]{3.5cm}
Model Poisoning
\begin{itemize}[align=parleft,left=0pt..1em]
    \item Backdoor Attack~\citep{bagdasaryan2020backdoor, chen2017targeted, wang2020attack, gu2017badnets, xie2019dba, fung2020limitations}
    \item Byzantine Attack~\citep{fang2020local, bhagoji2019analyzing, rong2022poisoning}
\end{itemize}
\end{minipage}}} & \multirow{3}{*}{\begin{minipage}[c]{3.5cm}\begin{itemize}[align=parleft,left=0pt..1em]
    \item Robust Aggregation~\citep{blanchard2017machine, guerraoui2018hidden, xie2018generalized, chen2017distributed, yin2018byzantine, li2019rsa, wu2020federated, pillutla2019robust, portnoy2022towards}
    \item Byzantine Detection~\citep{munoz2019byzantine, li2020learning, xie2020zeno++, fang2020local, zhang2022fldetector}
    \item Hybrid Mechanism~\citep{prakash2020mitigating, cao2021fltrust, han2020robust, zhao2022fedinv, zhu2019deep228}
\end{itemize}\end{minipage}} & \multicolumn{1}{c|}{\multirow{3}{*}{}}   & \multirow{3}{*}{} \\ \cline{3-3}
 &                            & \multicolumn{1}{c|}{\multirow{2}{*}{\begin{minipage}[c]{3cm}
Data Poisoning\citep{gu2017badnets, bagdasaryan2020backdoor, tolpegin2020data, fang2018poisoning, biggio2012poisoning, wang2020attack}
\begin{itemize}[align=parleft,left=0pt..1em]
    \item Label Flipping Attack~\citep{biggio2012poisoning}
    \item Poisoning Sample Attack~\citep{gu2017badnets, bagdasaryan2020backdoor}
\end{itemize}
\end{minipage}}} &     \begin{minipage}[c]{3.5cm}
\begin{itemize}[align=parleft,left=0pt..1em]
    \item Optimization-based Methods~\citep{li2020federated, karimireddy2020scaffold, wang2020tackling, reddi2020adaptive, DBLP:conf/iclr/AcarZNMWS21, reisizadeh2020robust}
    \item Knowledge-based Methods~\citep{jeong2018communication, seo2020federated, zhu2021data, li2019fedmd}
\end{itemize}
\end{minipage}              & \multicolumn{1}{c|}{}                                       &                   & \multicolumn{1}{c|}{ N/A}                                    &      N/A              \\
 &                            & \multicolumn{1}{c|}{}                                       &                   & \multicolumn{1}{c|}{}                                       &                  & \multicolumn{1}{c|}{}                                     &                  \\ 
\bottomrule
\end{tabular}}% 
\end{table*}

\subsection{Framework of Trustworthy Federated Learning}
% Federated learning provides a generalized way to explore distributed data. Unlike traditional centralized machine learning, FL has distinct characteristics (\eg, data isolation, communication, non-IID distribution, \etc) that distinguish FL trustworthiness practices from those in general AI and other learning types~\citep{liu2021trustworthy, zhang2022trustworthy} . To better illustrate our TFL framework, we first briefly describe the potential threats in the different stages of FL. 
Federated learning offers a more versatile approach to dealing with distributed data. It differs from traditional centralized machine learning in several ways, such as data isolation, communication, non-IID distribution, and others. These characteristics make FL unique and require different trustworthiness practices compared to general AI and other learning types~\citep{liu2021trustworthy, zhang2022trustworthy}. To provide a better understanding of our TFL framework, we will begin by briefly describing potential threats in the different stages of FL

\begin{itemize}
    \item \textbf{Data Processing.} 
    % In data processing, the threat comes from information leakage and malicious attack on the data. For example, malicious parties can conduct data poisoning attacks ~\citep{chen2017targeted, li2016data, alfeld2016data} on FL. \Eg, the parties can modify the labels of training samples with a specific class, so that the model performs badly in this class.
    In the data processing stage, threats arise from potential information leakage and malicious attacks on the data. For instance, malicious actors can conduct data poisoning attacks~\citep{chen2017targeted, li2016data, alfeld2016data} in FL. They may modify the labels of training samples with a specific class, resulting in poor performance of the model on this class.
    
    \item \textbf{Model Training.} 
    % During the training process, the malicious participant in FL can perform model poisoning attacks~\citep{xiao2010differential, xie2019dba, blanchard2017machine} to upload designed model parameters. The global model can have a very low accuracy due to the poisoned local updating. Besides model poisoning attacks, the Byzantine fault~\citep{blanchard2017machine} is also a common issue in distributed learning, where the parties may behave arbitrarily badly and upload random updates.
    During the model training process, a malicious participant in FL can perform model poisoning attacks~\citep{xiao2010differential, xie2019dba, blanchard2017machine} by uploading designed model parameters. The global model may have low accuracy due to the poisoned local updates. In addition to model poisoning attacks, Byzantine faults~\citep{blanchard2017machine} are also common issues in distributed learning, where the parties may behave arbitrarily and upload random updates.
    
    \item \textbf{Deployment and Inference.} 
    % If the learned model is published, inference attacks~\citep{shokri2017membership160,melis2019exploiting, nasr2019comprehensive129} can be conducted on it. The server can infer sensitive information about the training data from the exchanged model parameters. For example, membership inference attacks~\citep{shokri2017membership160, nasr2019comprehensive129} can infer whether a specific data record is used in the training. Note that the inference attacks may also be conducted in the learning process by the FL manager, who has access to the local updates of the parties.
    After the model is learned, inference attacks~\citep{shokri2017membership160, melis2019exploiting, nasr2019comprehensive129} can be conducted on it if it is published. The server may infer sensitive information about the training data from the exchanged model parameters. For example, membership inference attacks~\citep{shokri2017membership160, nasr2019comprehensive129} can infer whether a specific data record was used in the training. It is worth noting that inference attacks may also occur in the learning process by the FL manager, who has access to the local updates of the parties.
\end{itemize}

Then we briefly explain the three key core aspects of trustworthiness and its associated defense methods (Table~\ref{tab:taxonomy}).

\begin{itemize}
    \item \textbf{Privacy.} 
    % Privacy indicates how private data within FL can be protected to prevent it from being leaked. Specifically, the privacy of FL data and model parameters, which are regarded as confidential information belonging to their owners, should be guaranteed, and any unauthorized use of the data that can directly or indirectly identify a person or household should be prevented. These data cover a wide range of information, including name, age, gender, face image, fingerprints, \etc. Commitment to privacy protection is regarded as an important factor determining the trustworthiness of an FL system. This is not only because of the data value but also the requirement of the regulation and laws.
    Privacy refers to how private data within FL can be protected to prevent it from being leaked. It is crucial to guarantee the privacy of FL data and model parameters, which are considered confidential information belonging to their owners, and to prevent any unauthorized use of the data that can directly or indirectly identify a person or household. This data covers a wide range of information, including names, ages, genders, face images, fingerprints, etc. Commitment to privacy protection is an essential factor in determining the trustworthiness of an FL system. This is not only because of the data value but also because of regulatory and legal requirements.
    
    \item  \textbf{Robustness.} 
    % Robustness refers to the ability of FL to remain stable under any extreme condition, especially those that are created by attackers. This is important since real environments where FL systems are deployed are usually very complex and volatile. Robustness is an important factor affecting the performance of FL systems in empirical environments. The lack of robustness may also cause unintended or harmful behavior by the system, thus diminishing its trustworthiness.
    Robustness refers to the ability of FL to remain stable under extreme conditions, particularly those created by attackers. This is essential because real environments where FL systems are deployed are usually complex and volatile. Robustness is a vital factor that affects the performance of FL systems in empirical environments. The lack of robustness may also cause unintended or harmful behavior by the system, thereby diminishing its trustworthiness.

    \item  \textbf{Security.}  
    % guarantee the FL programs or system to be integrate and protect the correctness of the computation flow. Security refers to the protective measures taken to prevent unauthorized users in FL from gaining access to data.
% Security refers to the protective measures taken to prevent unauthorized users from gaining access to data. The core of security is to ensure that unauthorized users cannot access sensitive data. Accordingly, the security of FL is to ensure the confidentiality and correctness of computational data. Specifically, these data include training and prediction data, intermediate parameters, and the results of the trained model. Unlike privacy, which targets individuals, security focuses more on systems, institutions, and the public interest.
Security refers to the protective measures taken to prevent unauthorized users from gaining access to data. The core of security is to ensure that unauthorized users cannot access sensitive data. Therefore, the security of FL aims to ensure the confidentiality and correctness of computational data. Specifically, this data includes training and prediction data, intermediate parameters, and the results of the trained model. Unlike privacy, which targets individuals, security focuses more on systems, institutions, and the public interest.
% \vspace{-0.2cm}
\end{itemize}
% It is worth noting that previous FL works~\citep{yin2021comprehensive,yang2019federated} often refer to privacy and security indiscriminately and regard them as privacy-preserving federated learning (PPFL). However, these two concepts have different guarantees for TFL. We clarify the distinction between these two concepts. We consider security is the guarantee that supports the confidentiality and correctness of the system. The protection makes sure that attackers cannot have access to the internal TFL (\eg, eavesdropping on weight, gradients, and \etc).  While in TFL, privacy indicates how private information within FL can be protected to prevent it from being leaked. Some work points out that private information (\eg, data records, proprieties, membership) can be retrieved from the weight and gradient. Privacy guarantee makes sure, even if the attackers have access to the internal TFL (\eg, access to the weight, model, gradients \etc), the private information can still be preserved.
It is worth noting that previous works on FL~\citep{yin2021comprehensive, yang2019federated} often use the terms privacy and security indiscriminately and refer to them as privacy-preserving federated learning (PPFL). However, these two concepts have different guarantees for TFL, and we aim to clarify the distinction between them. In TFL, security guarantees the confidentiality and correctness of the system. Protection is necessary to ensure that attackers cannot access the internal TFL, such as weight, gradients, and other sensitive information. On the other hand, privacy in TFL indicates how private information within FL can be protected to prevent it from being leaked. Some studies have found that private information, such as data records, proprieties, and membership, can be retrieved from the weight and gradient. Thus, privacy guarantees that even if attackers can access the internal TFL, such as weight, model, gradients, \etc, private information can still be preserved.

Our survey focuses on providing technical solutions for each aspect of TFL in different development stages. This perspective sets it apart from recent related works, such as government guidelines~\citep{smuha2019eu_ai_307} that suggest building TFL systems through laws and regulations or reviews~\citep{brunet2019understanding_ai_52,smuha2019eu_ai_307} that discuss TFL realization from a high-level, non-technical perspective. Our main contributions are:

\begin{itemize}
\item We present a synopsis of threats and defense approaches for the core aspects of TFL (\ie, privacy, robustness, and security) in different development stages of FL (\ie, data processing, model training, and deployment) to provide a general picture of the field of Trustworthy Federated Learning.

\item We discuss the challenges of trustworthiness in FL, clarify existing gaps, identify open research problems, and indicate future research directions.
\end{itemize}

In the remaining sections, we organize the survey as follows: In \secref{sec:threats}, we provide an overview of existing threats in the TFL system to help readers understand the risks involved in building a TFL system from a software development perspective. In \secref{sec:security}, we detail the aspect of security, which ensures the confidentiality and correctness of computational data in TFL. In \secref{sec:robustness}, we detail the aspect of robustness, which makes a TFL system robust to the noisy perturbations of inputs and enables it to make trustworthy decisions. In \secref{sec:privacy}, we present the dimension of privacy, which guarantees a TFL system avoids leaking any private information.

\section{Threats in Trustworthy Federated Learning\label{sec:threats}}

Potential threats exist in all phases of federal learning, harming the trustworthiness of the system. The distribute nature of FL makes it vulnerable to information leakage and adversarial attack. In this section, we summarize and compare the existing studies on FL threats according to the aspects considered in \secref{sec:intro}.
\subsection{Threats in Data Processing of Federated Learning}

One main vulnerable stage of FL is the data processing. At this stage, raw data is cleaned, processed, and transformed into features that can be accepted by machine learning model. Due to the distributed nature of federated learning, Non-iid data and information leakage are common treats in unprotected FL environment. Meanwhile, malicious can to harm the model by sending poisoned data during data processing. In what follows, we summarize the potential threats in the data processing phase.

\vspace{0.1cm}\noindent\textbf{Information Leakage.} Information leakage exists in the data processing stages even if there is no direct data exchange in FL. In federal learning, although data is not directly involved in communication, the exchanged model parameters and gradients still contain private information. In the absence of data protection (\eg, encryption, perturbation, or anonymization), a direct exchange of model and gradient derived from the local data causes privacy leaks. A number of studies show that the raw record, membership, and properties can be inferred from weight and gradients~\cite{zhu2019deep228,zhao2020idlg_p_223}.

\vspace{0.1cm}\noindent\textbf{Non-IID Data.}
Due to distributed setting of FL, the data are preserved on isolated devices/institutions and cannot be centralized. Hence, the data samples are generated in various devices/institutions where the source distributions can be None Independent and Identically Distributed (None-IID) in many ways~\cite{hsieh2020non}. A number of studies~\cite{zhao2018federated, zhu2021federated} have indicated that the performance drop of FL in Non-IID settings is inevitable. The divergence of local updates~\cite{zhao2018federated} continues to accumulate, slowing down the model convergence and weakening model performance.

\vspace{0.1cm}\noindent\textbf{Data Poisoning Attack.}
% Data poisoning happen in the data pre-process procedure, the adversary can introduce a number of data sample it wishes to miss-classify with the desired target label into the training set. One common example is the label-flipping attack~\cite{biggio2012poisoning}. The labels of honest training examples of one class are flipped to another class while the features of the data are kept unchanged. For example, the malicious clients in the system can poison their dataset by flipping all 1s into 7s. A successful attack produces a model that is unable to correctly classify 1s and incorrectly predicts them to be 7s. Another weak but realistic attack scenario is sample poisoning attack~\cite{gu2017badnets}. Here, an adversary can modify individual features or small regions of the original training dataset to embed backdoors into the model, so that the model behaves according to the adversary’s objective if the input contains the backdoor features (\eg, a stamp on an image). However, the performance of the poisoned model on clean inputs is not affected. In this way, the attacks are harder to be detected. Data poisoning attacks can be carried out by any FL participant. The impact on the FL model depends on the extent to which participants in the system engage in the attacks, and the amount of training data being poisoned. Data poisoning is less effective in settings with fewer participants.
% 
% Data poisoning happen in the data pre-process procedure, the adversary can introduce a number of data sample it wishes to miss-classify with the desired target label into the training set.
Data poisoning occurs during the data pre-processing step. The adversary adds a number of data samples to the training set with goal of miss classify the target label.
% One common example is the label-flipping attack~\cite{biggio2012poisoning}. The labels of honest training examples of one class are flipped to another class while the features of the data are kept unchanged.
The label-flipping attack~\cite{biggio2012poisoning} is a common example. It does not change the data properties, but make the the target labels to another class.
% For example, the malicious clients in the system can poison their dataset by flipping all 1s into 7s. A successful attack produces a model that is unable to correctly classify 1s and incorrectly predicts them to be 7s.
For instance, hostile users can poison their data by replacing all labels of ``apple'' with labels of ``banana''. As a result, the trained classifier cannot correctly categorize ``apple'' and it will incorrectly expect ``apple'' to be ``banana''.
%  A successful attack produces a model that is unable to correctly classify 1s and incorrectly predicts them to be 7s. Another weak but realistic attack scenario is sample poisoning attack~\cite{gu2017badnets}.
\citet{gu2017badnets} design another practical poisoning attack called backdoor attacks.
%  Here, an adversary can modify individual features or small regions of the original training dataset to embed backdoors into the model, so that the model behaves according to the adversary’s objective if the input contains the backdoor features (\eg, a stamp on an image).
In this scenario, an attacker can alter specific features or subsections of the original training dataset to implant backdoors into the model, so that the model responds according to the adversary's intent if the input contains the backdoor features. 
% However, the performance of the poisoned model on clean inputs is not affected. In this way, the attacks are harder to be detected.
This kind of attack is difficult to identify as the performance of the poisoned model with clean inputs remains mostly unchanged. 
Note that any FL clients are capable of launching a data poisoning attack.
% The impact on the FL model depends on the extent to which participants in the system engage in the attacks, and the amount of training data being poisoned. Data poisoning is less effective in settings with fewer participants.
The impact on the FL model is dependent on the frequency of the engagement of clients in attacks and the quantity of tainted training data.

\subsection{Threats in Model Training of Federated Learning}

\vspace{0.1cm}\noindent\textbf{Model Poisoning Attack.}
% Model poisoning attacks aim to poison local model updates before sending them to the server or insert hidden backdoors into the global model~\cite{bagdasaryan2020backdoor}. due to the frequently model update, the model poisoning attacks are consider to be more effective than the data poisoning attack. Depending on the attack objectives, poisoning-based attacks can be either untargeted attacks or targeted attacks. In targeted model poisoning, the adversary’s objective is to cause the FL model to miss-classify a set of chosen inputs with high confidence. The poisoned updates can be adversarial manipulations of the training process~\cite{blanchard2017machine} or be generated by inserting hidden backdoors, and even a single-shot attack may be enough to introduce a backdoor into a model~\cite{bagdasaryan2020backdoor}. In untargeted attacks poisoning, the adversary’s objective is to reduce the performance of FL model. For example in byzantine attack, The malicious participants tailor its outputs to have similar distribution as the correct model updates to make itself hard to detect. it behave completely arbitrarily to decrease the model performance.
% Model poisoning attacks aim to poison local model updates before sending them to the server or insert hidden backdoors into the global model~\cite{bagdasaryan2020backdoor}.
Model poisoning attacks are intended to poison local updates before they are communicated to the server or to implant hidden backdoors inside the global model~\cite{bagdasaryan2020backdoor}. 
%  due to the frequently model update, the model poisoning attacks are consider to be more effective than the data poisoning attack.
Because models are continuously updated, it is believed that model poisoning attacks more effective compared to data poisoning attacks. 
% Depending on the attack objectives, poisoning-based attacks can be either untargeted attacks or targeted attacks.
Depending on the objectives of the attack, there are two types of poisoning-based attacks (\ie, untargeted and targeted attacks).
%  In untargeted attacks poisoning, the adversary’s objective is to reduce the performance of FL model. For example in byzantine attack, The malicious participants tailor its outputs to have similar distribution as the correct model updates to make itself hard to detect. it behave completely arbitrarily to decrease the model performance.
In targeted model poisoning, the adversary seeks to induce the FL model to misclassify a collection of specified inputs with high probability.
The poisoned updates can be injected via manipulating the training process or generated by incorporating hidden backdoors~\cite{blanchard2017machine}. \citet{bagdasaryan2020backdoor} shows the simple single-shot attack could be sufficient to harm the FL model. In untargeted attack poisoning, the adversary seeks to undermine the performance of the FL model. In a byzantine attack~\cite{fang2020local, bhagoji2019analyzing, rong2022poisoning}, the adversary players change their outputs to have the same distribution as the right model updates in order to evade detection. It behaves completely arbitrarily to diminish model performance.

\vspace{0.1cm}\noindent\textbf{Privacy Leakage and Communication Eavesdropping.}
Information leakage and communication eavesdropping are common issue during the training process, especially during weight and gradient updates.
%  Note that Information Leakage is threat of privacy while communication eavesdropping is threat of security.
It is important to note that information leakage is a privacy threat, whereas communication eavesdropping is a security threat.
% In FL, three types of important data need to be transmitted between the participants and the aggregator: the local weight/gradient, an aggregated weight/gradient, and the final model, all of which contain the necessary information that can be eavesdropped and further utilized to reveal sensitive information about the training datasets~\cite{fredrikson2015model53,yu2019parallel_p_212}.
In FL, three types of data must be transferred between clinets and server: weight, gradients, and final model. 
Each of these data formats contains sensitive information about the training datasets that can be intercepted and utilized to reveal sensitive information~\cite{fredrikson2015model53,yu2019parallel_p_212}. 
Clients communicate gradients/weights to the server, which collects them and returns them to clients for model updating in gradient/weight-update-based FL systems~\cite{lu2020decentralized_p_115}.
% In deep network models, the gradients are usually calculated by back-propagating the loss of the training datasets through the entire network.
Gradients are commonly calculated in deep neural network models by back-propagating throughout the whole network.
% Specifically, the gradient of a layer is calculated using the current layer’s data features and the error from the upper layer. Similarly, the local model weight is calculated based on the participant’s local dataset. Therefore, they (weight update, gradient update and the final model) contain sensitive information of local data ~\cite{su2019securing_p_167, qu2020decentralized_p_142}.
Specifically, the gradient is compute via the current layer activation and the error propagate fromt the loss. in the same way, the local model weight is update using  the gradients compute form the local dataset. As a result, weight updating, gradient updating, and final model) may contain sensitive details about local data~\cite{su2019securing_p_167, qu2020decentralized_p_142}.

\subsection{Threats in Development of Federated Learning}

\vspace{0.1cm}\noindent\textbf{Inference Phase Data Security.}
In the phase of deployment or inference, the task initiator usually deploys the carefully-trained model to the cloud server to provide prediction services and get the related payment. The model parameters, as the intellectual property of the model owner, need to be effectively protected in order to continuously generate value. However, the curious cloud server may steal the model parameters during the deployment process. 
Furthermore, to accomplish the prediction task, users need to upload their data, which may contain sensitive information such as gender, age, health status, \etc being exposed to the cloud server. A lot of work~\cite{hesamifard2017cryptodl, gilad2016cryptonets,fenner2020privacy,chen2022x,rouhani2018deepsecure,liu2017oblivious,riazi2018chameleon,juvekar2018gazelle, mishra2020delphi, riazi2019xonn,chaudhari2019astra, rathee2020cryptflow2,huang2022cheetah,zhu2022securebinn,rathee2021sirnn} emerged in order to address data leakage in the inference phase, and we will describe these studies in detail afterwards. 

\vspace{0.1cm}\noindent\textbf{Inference Phase Attack.}
The inference phase focuses on determining how to provide the query service to consumers, and it is also vulnerable to inference attack~\cite{shokri2017membership, melis2019exploiting}.
% These attacks are carried out in the inference phase when the model has been trained. They are called evasion or exploratory attacks~\cite{}.
This kind of attack is launched in the inference phase after the model has been trained, which is usually referred to as an evasive or exploratory attack. 
%  Generally, the objective is not to modify the trained model, but to produce wrong predictions or to collect information about the characteristics of the model. During the inference phase, the risk of a privacy leakage is mainly related to the final model, which is released to the participants or is provided as a service platform.
In general, the purpose of an inference attack is to produce inaccurate predictions or collect information about the model's attributes rather than to alert the trained model. The threat during the inference phase is mostly associated with the final model, which is either published to clients or provided as an API for external users. 
% here are two main risks of privacy leakage: (1) attacks based on model parameters and (2) attacks based on model queries.
There are two key threats associated with inference attacks: (1) Model-based attacks and (2) Query-based attacks. Attackers have access to the model parameters and thus the query results as well as any intermediate computations, they can extract sensitive information about the participants' training datasets~\cite{fredrikson2015model53,zhang2012functional}.

\section{Security~\label{sec:security}}

In this section, TFL is introduced from a security perspective, and we refer to these studies as Secure Federation Learning~(SFL). Specifically, SFL guarantee the confidentiality and integrity of data and model during FL training and inference by employing secure computing include Secure Multi-Party Computation~(SMPC)~\citep{yao1982protocols,Yao1986MPC} and Homomorphic Encryption~(HE)~\citep{paillier1999public, elgamal1985public, abney2002bootstrapping}. In~\secref{sec:definition of sfl}, We first analyze the threats that exist in SFL and give a formal definition of SFL based on this.
Subsequently the secure computing techniques used to address the threats present in SFL are presented in~\secref{sec: defense methods}. Finally, we provide an overview of SFL research works in terms of data security, model security, and system security in~\secref{sec:SFL}.

% The roadmap of this section is as follows. In~\secref{sec: thread model}, we introduce the threat model of the SFL algorithms. In~\secref{sec: defense methods}, we present the defense methods used in the SFL algorithms. In~\secref{sec:SFL}, we provide an overview of SFL research works in terms of data security, model security, and system security. Finally, in~\secref{sec:SFL summary} we summarize the main challenges of SFL and suggest some valuable research directions. 

% \begin{figure}
%     \centering
%     \includegraphics[width=\textwidth]{Content/security/pictures/SFL-Category.jpg}
%     \caption{\textcolor{red}{Revise: Categories of Secure Federated Learning}. ~\textcolor{blue}{IK: Need to update the taxonomy and the caption.} }
%     \label{fig:SFL-Arc}
% \end{figure}

%***************************************************************************
% 给出SFL的定义以及SFL所考虑的安全模型。
\subsection{The Definition of Secure Federated Learning}
\label{sec:definition of sfl}
In this section, we describe the threat models present in the FL training and prediction process and give the definition of SFL based on these threat models. Specifically, there are different threats to FL in different scenarios. These threats can be abstracted into different threat models. An SFL algorithm is secure under a certain threat model means that it can resist all attacks under that threat model.

The threat model of SFL can be divided into a semi-honest~(passive) model and a malicious~(active) model according to the system's tolerance to the adversary's capabilities. Specifically, under the assumption of a semi-honest model, the adversary will perform operations according to the protocol but tries to mine the private data of other participants through the information obtained in the process of executing the SFL protocol. And the security of SFL under the semi-honest model requires that the user's private information is not available to the adversary. Unlike the semi-honest model, in the malicious model the adversary will violate the protocol in order to obtain private data.
The security of SFL under the malicious model also requires that the user's private data is not available to the adversary. In addition, the security strengths of the SFL algorithms, sorted from weak to strong, are: abort, fair, and guaranteed output delivery (GOD). In detail, abort means that the security detects malicious behavior and terminates the protocol. Fairness means that the dishonest participant can get the output when and only when the honest participant gets the output result in the secure computing protocol. The GOD means that the dishonest participant cannot prevent the honest participant from obtaining the output during protocol execution.
Furthermore, the threat model of SFL can be divided into honest-majority and dishonest-majority according to the percentage of participants controlled by the adversary in the system. Specifically, the number of adversary-controlled participants in the honest majority model is less than half of the total number of participants. In contrast, in the dishonest majority model, the number of participants controlled by the adversary is greater than or equal to half of the total number of participants. % 给出SFL的形式化定义
Based on the threat model of SFL, we give the following definition of SFL. 

\begin{definition}
The training process of FL can be seen as a function of  $m$-ary functionality,  denoted by 
\begin{equation}
f:(\{0, 1\}^{*})m \rightarrow (\{0, 1\}^{*})m.
\end{equation}
Specifically, $f$ is a random process mapping string sequences of the form $x = (x_1,\dots, x_m)$ into sequences of random variables, $f_1(x), \dots , f_m(x)$ such that, for every $i$, the $i$-th party $P_i$ who initially holds an input $x_i$, wishes to obtain the $i$-th element in $f (x_1, \dots , x_m)$ which is denoted by $f_i(x_1, \dots, x_m)$.  The inference process of FL can be seen as a function of  one-ary functionality,  denoted by 
\begin{equation}
g : \{0, 1\}^{*} \rightarrow \{0, 1\}^{*}.
\end{equation}
Specifically, $g$ is a random process mapping string sequences of the form $y$ into sequences of random variables, $g(y)$, such that the client $C$ who initially holds an input $y$ wishes to obtain the $g(y)$. We call an FL algorithm as an SFL algorithm if it can complete the calculation of the training or inference process under a given threat model.
\end{definition}
%***************************************************************
% 介绍不同的安全防御方法，主要包括SMPC、HE和TEE。其中SMPC和HE均属于安全计算技术，而TEE则属于硬件隔离技术。
\subsection{Defense Methods in Secure Federate Learning}
\label{sec: defense methods}
In this section, we introduce the  defense techniques commonly used in the SFL algorithms, such as SMPC, HE, and Trusted Execution Environment~(TEE). Accordingly, SFL algorithms using different secure computing techniques will have different characteristics. For example, the SMPC-based secure computing algorithm can theoretically realize any computation, but it needs to spend a lot of communication costs. Conversely, the HE-based secure computing algorithm can implement addition and multiplication operations without any communication costs, but it requires huge computational overhead. 

\subsubsection{Secure Multi-Party Computation}
Secure Multi-Party Computation~(SMPC) originated from the millionaire problem proposed in~\citep{yao1982protocols}.  The goal of SMPC is to allow a group of mutually untrusted data owners to work together on the computation of a function under the condition that the confidentiality of their independent data is not compromised. The main techniques currently implementing the SMPC protocol include Garbled Circuit~(GC)~\citep{Yao1986MPC}, Oblivious Transfer~(OT)~\citep{rabin2005exchange}, and Secret Sharing~(SS)~\citep{shamir1979share}. 
All these techniques have limitations and usually require a combination with other techniques to construct efficient SFL algorithms. For example, in spite of, GC can theoretically enable secure computation of arbitrary functions in constant rounds, but transmitting the complete encrypted circuit will result in high communication costs. In contrast, while SS can achieve secure computation at a lower communication cost, it requires a larger number of communication rounds for complex operations.

\vspace{0.1cm}\noindent\textbf{Oblivious Transfer.} Oblivious transfer~(OT) is proposed in~\citep{rabin2005exchange}. As a very important primitive in cryptography, OT not only can implement the SMPC protocol independently but also can integrate with other technologies to complete the construction of the SMPC protocol. There are generally two parties in the OT protocol namely, the \emph{sender} and the \emph{receiver}. The goal of the OT protocol is to enable the receiver to obtain certain information from the sender obliviously on the premise that the sender and receiver's respective private information is not leaked.

\vspace{0.1cm}\noindent\textbf{Garbled Circuit.} Garbled circuit~(GC) is a two-party SMPC protocol proposed in~\citep{yao1982protocols}. The two parties in the GC are called \emph{garbler} and \emph{evaluator}. Suppose the input information of the garbler is $x$, and the input information of the evaluator is $y$. The main idea of GC is to convert computational functions into boolean circuits. In the obfuscation stage, the garbler converts the Boolean circuit corresponding to the calculation function into a garbled circuit and sends the garbled circuit and the random input label corresponding to $x$ to the evaluator. 
The evaluator executes the OT protocol by interacting with the garbler and obtains the corresponding random input label of $y$. The evaluator decrypts the garbled circuit using the random input tag to get the calculation result.
Finally, the evaluator sends the decrypted calculation result to the garbler. Since the garbled circuit and the random input label of $x$ are random values for the calculator, they do not contain any information of $x$, and the security of the OT protocol used ensures that the information of $y$ will not be leaked to the obfuscator. Therefore, the obfuscation circuit ensures that both parties involved in the calculation can obtain the calculation result without revealing their respective input data.

\vspace{0.1cm}\noindent\textbf{Secret Sharing.} Secret sharing~(SS) is a technique independently proposed by Shamir~\citep{shamir1979share} and Blackly~\citep{blakley1979safeguarding} with its full name called $(t, n)$-threshold secret sharing schemes, where $n$ is the number of parties and $t$ is a threshold value.
The security of SS requires that any less than $t$ parties cannot obtain any secret information jointly.
As a special case of secret sharing, $(2,2)$-\emph{additive} secret sharing contains two algorithms: $Shr(\cdot)$ and $Rec(\cdot, \cdot)$.
Let $([u]_0, [u]_1)$ be the additive share of any $u$ on $Z_L$. $Shr(u) \rightarrow ([u]_0, [u]_1)$ is used to generate the share by randomly selecting a number $r$ from $Z_L$, letting $[u]_0=r$, and computing $[u]_1=(u - r)\mod L$. Note that due to the randomness of $r$, neither a single $[u]_0$ nor $[u]_1$ can be used to infer the original value of $u$. The algorithm $Rec([u]_0, [u]_1)\rightarrow u$ is used to reconstruct the original value from the additive shares, which can be done by simply calculating $([u]_0 + [u]_1) \mod L$.  The additive secret-sharing technique has been widely used to construct SMPC protocols for ML operations~\citep{mohassel2017secureml,Wagh2019SecureNN,wagh2020falcon,ryffel2020ariann}. GMW~\citep{Goldreich2019GMW} represents a function as a Boolean circuit and uses the value of XOR-based SS. Compared with addition secret sharing, "XOR" is used instead of addition and "AND" is used instead of multiplication in GMW.
%Let's take the multiplication $u \cdot v$ as an example:
%Let ${P_j}$ with $j \in \{0,1\}$ be two parties that are used to execute the SMPC protocol. Each party $P_j$ will be given one additive share $([u]_j, [v]_j)$ of the operation inputs for $j \in \{0,1\}$. Then, the additive shares of $u\cdot v$ can be computed with Beaver-triples ~\citep{Beaver1991triples}: $(a,b,c)$ where $a, b \in Z_{L}$ are randomly sampled from $Z_{L}$ and $c = a \cdot b$.Specifically, for each $j \in \{0, 1\}$,  $P_j$ first calculates $[d]_j = [u]_j-[a]_j$ and $[e]_j = [v]_j - [b]_j$. Then, they send the $[d]_j$ and $[e]_j$ to each other and reconstruct $d= Rec([d]_0, [d]_1) = [d]_0+[d]_1, e =  Rec([e]_0, [e]_1) = [e]_0+[e]_1$. Finally, the additive share of $u \cdot v$ can be computed using $[u\cdot v]_j = -jd \cdot e +[y]_j \cdot e + d \cdot [v]_j + [c]_j$. Note that there will be $1$ round of two-way communication between the two parties for exchanging $[d]_j$ and $[e]_j$ in one SS-based private-preserving multiplication.
% 
%Unfortunately, there exist many operations (e.g., division, exponentiation, \emph{etc}\onedot.) that cannot be constructed using purely additive secret sharing on $Z_L$. Some approximation methods such as Newton-Raphson iteration method and Taylor expansion have been exploited for designing additive SS-based protocols of these operations. Details of the approximation methods and other SMPC protocols can be found in ~\citep{knott2021crypten}. 
\subsubsection{Homomorphic Encryption}
Homomorphic encryption~(HE) makes the operation of plaintext and ciphertext satisfy the homomorphic property, \emph{i.e}\onedot, it supports the operation of ciphertext on multiple data, and the result of decryption is the same as the result of the operation of the plaintext of data. Formally, we have 
% \begin{equation}\label{eq:HE1}
%     \forall x \in \mathcal{X}, x_1, x_2, \cdots, x_n \rightarrow [x_1], [x_2], \cdots, [x_n] , 
% \end{equation}
% we have 
% \begin{equation}\label{eq:HE2}
% f([x_1], [x_2], \cdots, [x_n]) \rightarrow [f(x_1, x_2, \cdots, x_n)].
% \end{equation}
\begin{equation}\label{eq:HE1}
    f([x_1], [x_2], \cdots, [x_n]) \rightarrow [f(x_1, x_2, \cdots, x_n)], \text{ where }  \forall x \in \mathcal{X}, x_1, x_2, \cdots, x_n \rightarrow [x_1], [x_2], \cdots, [x_n].
\end{equation}
Homomorphic encryption originated in 1978 when ~\citet{rivest1978data}  proposed the concept of privacy homomorphism. However, as an open problem, it was not until 2009, when Gentry proposed the first fully homomorphic encryption scheme~\citep{gentry2009fully} that the feasibility of computing any function on encrypted data was demonstrated. According to the type and number of ciphertext operations that can be supported, homomorphic encryption can be classified as partial homomorphic encryption~(PHE), somewhat homomorphic encryption~(SHE), and fully homomorphic encryption~(FHE). Specifically, PHE supports only a single type of ciphertext homomorphic operation, mainly including additive homomorphic encryption~(AHE) and multiplicative homomorphic encryption~(MHE), represented by Paillier~\citep{paillier1999public}, and ElGamal~\citep{elgamal1985public}, respectively. SHE supports infinite addition and at least one multiplication operation in the ciphertext space and can be converted into a fully homomorphic encryption scheme using bootstrapping~\citep{abney2002bootstrapping} technique. The construction of FHE follows Gentry's blueprint, \emph{i.e}\onedot, it can perform any number of addition and multiplication operations in the ciphertext space. Most of the current mainstream FHE schemes are constructed based on the lattice difficulty problem, and the representative schemes include BGV~\citep{brakerski2014leveled}, BFV~\citep{brakerski2012fully,fan2012somewhat}, GSW~\citep{gentry2013homomorphic}, CGGI~\citep{chillotti2016faster}, CKKS~\citep{cheon2017homomorphic}, \emph{etc}\onedot

% This encryption scheme is an effective approach to protect confidentiality in standard federated learning process, where exchange model parameters or model updates during training. This technique has been widely used in a lot of FL works ~\citep{asad2020fedopt, chen2020fedhealth, dong2020eastfly, hao2019efficient, hardy2017private, aono2017privacy, zhang2020batchcrypt, zhang2020privacy, zhao2020smss}.

\subsubsection{Trusted Execution Environment}
A Trusted Execution Environment~(TEE)~\citep{subramanyan2017formal} enables a certain part of the federated learning process into a trusted environment in the cloud, whose code can be attested and verified. It provides several properties which guarantee that codes could be executed faithfully and privately.  In detail, the confidentiality of TEEs guarantees the program process execution is secret, and the state of code is private; while the integrity of TEE ensures that the code's execution cannot be affected. In addition, the measurement of TEE provides the remote party with proof of the code being executed and its starting state.

The main function of TEE in SFL is to reduce the attack surface of adversaries. For external adversaries, TEE prevents them from stealing training data and intermediate results of the model training process. For internal adversaries, such as servers and participants, TEE can prevent collusion attacks, model reversal attacks, backdoor attacks, \emph{etc}\onedot, between them. Furthermore, TEE can be used to protect the model parameter information. 
%The problems with TEE in SFL applications include 1). TEE memory capacity is too small making it difficult to train larger models. 2). TEE is vulnerable to channel measurement attacks, leading to data leakage. 

%***************************************************************
% 介绍当前SFL的相关工作。首先介绍当前的分类方式。
% 按阶段分为安全训练和安全预测两类；
% 按是否采用明密文训练非为部分安全和完全安全两类；
% 按照所采用的安全计算技术分为四类。
\subsection{Secure Federated Learning Works}
\label{sec:SFL}
In this section, we categorize the SFL-related works in terms of data security, model security, and system security. The specific roadmap is as follows. First, in data security, we give an overview of SFL algorithms that only protect data security. This type of algorithm is generally applied in the training phase.
To balance security and efficiency, they protect only part of the training process and we refer to it as partially secure federated training~(PSFT). 
Next, in model security, we review SFL algorithms that protect both model and data security. These algorithms enable fully secure federated training~(FSFT) while supporting secure federated inference~(SFI). Finally, we review representative security systems in SFL.

% We will categorize the current work related to SFL in several dimensions, such as the stage of defense technology application, the type of defense technology and the security strength of the algorithm.
%First, SFL algorithms are classified into secure federated training (SFT) algorithms and secure federated inference (SFI) algorithms depending on the stage in which the secure computing technique is applied.
%Secondly, SFL algorithms are classified into partially secure federated algorithms and fully secure federated algorithms according to whether the strategy of mixed training with plain texts is used. Finally, SFL algorithms are classified into four categories \emph{i.e}\onedot, SFL-based-SMPC, SFL-based-HE, SFL-based-TEE and SFL-based-Hybrid, according to the different secure computing technologies used. It is worth noting that there are inclusion relationships between these divisions, e.g., locally secure federal algorithms exist only in secure federal training algorithms.  Through the analysis of the current state-of-the-art work of each type of SFL, we summarize the advantages and disadvantages of each type of scheme, and based on their characteristics, we give the possible applicable scenarios of each type of scheme in reality. 

% ***************************************************************************

% 介绍安全联邦训练算法的相关工作。
\subsubsection{Data Security}
\label{subsec:data security}
\begin{figure*}
    \centering
    \includegraphics[scale=0.545]{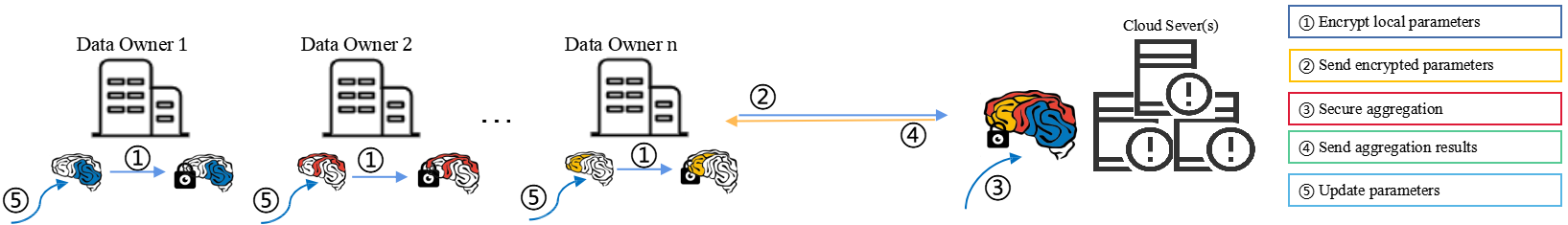}
    \caption{The pipeline of partially secure federated training.}
    \label{fig:PFLT}
    \vspace{-2.0em}
\end{figure*}

%In this section, we introduce the related work of SFT. Depending on whether a structured protection approach with a mixture of plaintext and ciphertext training is used, we classify secure training federated learning algorithms into two categories, namely, partially secure federation learning (PSFT) algorithms and fully secure federation learning (FSFT) algorithms. 
Here, we present an overview of the SFL algorithms that only protects data security, \emph{i.e}\onedot, PSFT. Specifically, We categorize the PSFT algorithms according to the different defense techniques used in them. As a class of SFL algorithms, PSFT protects part of the federal training process. In detail, the PSFT algorithm completes the training process by protecting the user's local model gradients. The pipeline of PSFT is shown in \figref{fig:PFLT}. 
% A summary of the PSFT algorithm is shown in~\tabref{table: PSFT}.

\vspace{0.1cm}
\noindent\textbf{SMPC-based PSFT.}
The SMPC-based PSFT algorithms have the advantage of low computational cost and communication volume. However, they usually requires more communication rounds and leaks the aggregation parameters to the server. As a well-known work,  ~\citet{bonawitz2017practical} proposes the first SMPC-based PSFT algorithm by using the \textit{Diffie-Hellman} key exchange protocol ~\citep{diffie2022new} and secret sharing. Specifically, it can protect the local parameters under the semi-honest (malicious) model by three(four) rounds of communication, when there are participants dropped in each round. On top of that, ~\citet{mandal2018nike} reducing the communication cost by using non-interactive key generation and L-regular graphs. Subsequently,~\citet{SecAgg} demonstrated that each client only needs to share the public key and encrypted share with some of the clients to accomplish secure aggregation. Based on this finding, they further reduce the communication and computation costs. In addition to this, there are some SMPC-based PSFT algorithms~\citep{dong2020eastfly, beguier2020safer} that use multiple non-colluding aggregation servers to protect local parameters. Specifically, participants send the shares of local parameters to the corresponding aggregation servers to achieve secure aggregation. Unlike ~\citep{bonawitz2017practical} which considers user dropouts, this type of work focuses on how to reduce the amount of communication transmitted by participants. As the first work, ~\citet{dong2020eastfly} proposes to reduce the communication volume by using a quantization approach. Inspired by this, ~\citet{beguier2020safer} further reduces the communication and computational cost of the PSFT algorithm by fusing techniques such as model sparsification, model quantization, and error compensation, making the communication of the PSFT algorithm comparable to or lower than FedAvg.\\
  
%******************************************************
\vspace{0.1cm}\noindent\textbf{HE-based PSFT.} 
%HE technique is widely used in the parameter fusion phase in SFL~\citep{aono2017privacy, zhang2020batchcrypt, zhang2020privacy, hao2019efficient, liu2019secure} because it can directly perform the required computation on the ciphertext for the plaintext. 
Compared with the SMPC-based PSFL algorithms, the HE-based PSFT algorithms have the advantage of additionally ensuring that the global model parameters are not leaked to the curious server.  This advantage further reduces the risk of data leakage during the model training phase. Although, the HE-based PSFT algorithms provides strong security guarantees but adds significant computational and communication overheads. Specifically, directly using HE algorithms, such as Paillier cryptosystem to implement security parameter aggregation, causes its computational cost to account for 80\% and increases communication by more than 150 times~\citep{zhang2020batchcrypt}. To reduce the computation and communication overhead of the HE-based PSFT algorithms, ~\citet{aono2017privacy} proposes a packetized computation, which effectively improves the computation and communication efficiency by encrypting multiple plaintexts after encoding them into one large integer while guaranteeing the correctness of ciphertext computation. Subsequently, ~\citet{zhang2020batchcrypt} further reduces the computation and communication overhead of HE-based PSFT algorithms by encrypting the quantized gradients after packing them by deflating, cutting, and quantizing them in advance.  In addition, ~\citep{xu2019verifynet_p_198, guo2020v, zhang2020privacy} designs HE-based PSFT algorithms with verifiable features by introducing bilinear aggregation signatures, homomorphic hashing, ~\emph{etc}\onedot, which effectively prevents malicious servers from corrupting or forging aggregation results.

\vspace{0.1cm}\noindent\textbf{TEE-based PSFT.} Unlike SMPC and HE, which achieve security assurance for FL at the algorithm level, TEE uses hardware isolation to reduce the risk of data leakage~\citep{zhang2021citadel, cheng2021separation, chen2020training, mo2019efficient}. Specially, the TEE effectively reducing the attack surface of adversaries in the FL system as well as preventing collusive attacks by participants in the FL system. The main problems faced by TEE in the construction of SFL systems include the lack of storage space and vulnerability to side-channel attacks. To address these problems, ~\citet{cheng2021separation} adopts the "Separation of power" approach to alleviate the problem of insufficient TEE memory by using multiple TEEs as aggregation servers to complete the aggregation of model parameters. At the same time, participants shuffle the parameters by random permutation before uploading them, effectively preventing TEE from side-channel attacks. In addition, ~\citet{chen2020training} uses TEE to protect the integrity of the FL training process and prevent malicious participants from tampering with or delaying the local training process. Furthermore, TEE can also provide additional model protection for FL algorithms, making the trained model accessible only to the task initiator after the FL task ends~\citep{zhang2021citadel}. 

In addition to this, some PSFT algorithms consider scenarios in which participants' data are divided according to features. In this setting,~\citep{vfl_smpc_1, vfl_smpc_2} design PSFT algorithms about linear regression and xgboost based on SMPC.~\citep{vfl_he_1,vfl_he_2,vfl_he_3,vfl_he_4,vfl_he_5,vfl_he_6,vfl_he_7,vfl_he_8} consider the design of the PSFT algorithms by using HE.
Specifically,~\citep{vfl_he_1,vfl_he_6} propose PSFT algorithms on logistic regression.~\citep{vfl_he_2,vfl_he_3,vfl_he_4} propose the PSFT algorithm about tree-based model.~\citep{vfl_he_5,vfl_he_7,vfl_he_8} design PSFT algorithms about neural network. Furthermore,~\citet{vfl_tee} uses TEE to design PSFT algorithm.~\citep{vfl_he_smpc_1,vfl_he_smpc_2,vfl_he_smpc_3,vfl_he_smpc_4} further improve the security of PSFT algorithms by fusing SMPC and HE. Recently,~\citet{vfl} provides an overview of federal learning algorithms by features.

In general, PSFT algorithms based on different technologies have their own characteristics. Specifically, the SMPC-based PSFT algorithms have good computational efficiency and can solve the user dropout problem using SS, making them more suitable for scenarios where the participants are mobile devices. In contrast, the HE-based PSFT algorithm is more suitable for scenarios where participants have stable communication and strong computational power. Furtermore, TEE can provide stronger hardware protection on top of the algorithm protection. Although, the good efficiency of PSFT makes it possible to train complex models, however, PSFT leaks global parameters to users or aggregation servers. This may compromise the confidentiality of the data to some extent. 
%*****************************************
% \vspace{0.1cm}
% \noindent\textbf{Hybrid-based PSFT.} In addition to using one secure computing technique, there are several works that further improve the security of PSFT by combining multiple secure computing techniques.  For example, ~\citet{xu2019verifynet} protects local model parameters by using HE and prevents participants from dropping out during training by using SS techniques. In addition ~\citet{zhao2020smss} constructs "SMSS" by using SS and HE, which can achieve the selection of federal participants without revealing the user's private data.

\begin{figure}
    \centering
    \includegraphics[scale=0.8]{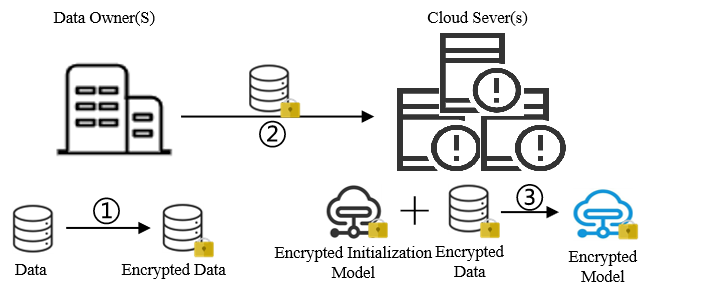}
    \caption{The pipeline of fully secure federated training.}
    \label{fig:FSFT}
    \vspace{-1.5em}
\end{figure}

\subsubsection{Model Security}
\label{subsec:model security}

In the following, we present an overview of SFL algorithms that can protect both model and data security. %, namely, FSFT and SFI.

\vspace{0.1cm}
\noindent\textbf{Fully Secure Federated Learning.}
Compared to the PSFT algorithms, which can only provide partial protection, FSFT algorithms achieve complete protection of the federation training process. Specifically, they convert the basic operations in machine learning such as matrix multiplication, and activation functions into corresponding secure operations using secure computing techniques. The FSFT algorithms usually adopts an architecture of outsourced computing, whose security relies on the assumption of non-collusion among the outsourced computing servers. 
In detail, the data and model owner first encrypts the data and model and sends it to the cloud servers. Then, the cloud servers the use the encrypted data and model interactions to complete the FL training process to obtain the encrypted model parameters. The pipeline of FSFT is shown in~\figref{fig:FSFT}. We classify the work of FSFT according to the number of servers. A summary of the FSFT algorithms is shown in~\tabref{table: FSFLT}. 

\begin{table*}[t]
  \caption{Fully Secure Federated Train Algorithms}
  \label{table: FSFLT}
\resizebox{\textwidth}{!}{
\begin{threeparttable}
\begin{tabular}{*{7}{c}}
  \toprule
   \multirow{2}*{No. of Servers} & \multicolumn{2}{c}{Security Model} & \multirow{2}*{Secure Defense Methods} &  \multirow{2}*{Model} & \multirow{2}*{Dataset} & \multirow{2}*{References}\\
  \cmidrule(lr){2-3}
  & Semi-honest & Malicious \\
  \midrule
  $2$ & D  & \ding{56} & GC/OT/SS/HE &  FC~\citep{mohassel2017secureml} & MNIST &SecureML  ~\citep{mohassel2017secureml}  \\
  $2$ & D  & \ding{56}  & GC/OT/SS &  FC~\citep{agrawal2019quotient} & MNIST &QUOTINENT~\citep{agrawal2019quotient}  \\
  $2$ & D  & \ding{56}   & GC/OT/SS/HE &  FC~\citep{patra2021aby2}  & MNIST & ABY2.0~\citep{patra2021aby2}  \\
  $2$ & D  & \ding{56} & GC/OT/SS/HE &  PR~\citep{kelkar2022secure} & Smoking and Cancer & \citet{kelkar2022secure}\\
  $3$ & H  & Abort & GC/OT/SS &  FC~\citep{Mohassel2018ABY3} & MNIST & ABY3~\citep{Mohassel2018ABY3}\\
  $3/4$ & H  & \ding{56}  & SS &  LeNet\citep{Wagh2019SecureNN} & MNIST & SecureNN~\citep{Wagh2019SecureNN}\\
  $3$ & H  & Abort & SS &  VGG16~\citep{wagh2020falcon} & Tiny ImageNet & FALCON~\citep{wagh2020falcon}\\
 $3$ & H  & \ding{56} & SS &  VGG16~\citep{ryffel2020ariann} & Tiny ImageNet & ARIANN~\citep{ryffel2020ariann} \\
 $3$ & H  & Fairness & SS &  FC~\citep{patra2020blaze} & Parkinson &  BLAZE~\citep{patra2020blaze} \\
  % $3$ & H  & GOD & SS &  VGG16~\citep{watson2022piranha} & CIFAR-10 & Piranha~\citep{watson2022piranha}  \\ 
 % $2/3/4$ & H  & - & SS &  VGG16~\citep{tan2021cryptgpu} & Tiny ImageNet~\citep{tan2021cryptgpu} & CryptGPU~\citep{tan2021cryptgpu}\\ 
  $4$ & H  & Fairness & GC/SS &  FC~\citep{li2019privpy} & MNIST & PrivPy~\citep{li2019privpy}\\
  $4$ & H  & \ding{56}  & SS &  FC~\citep{chaudhari2019trident} & MNIST & Trident~\citep{chaudhari2019trident} \\
 $4$ & H  & GOD & SS &  FC~\citep{dalskov2021fantastic} & MNIST & Fantastic Four~\citep{dalskov2021fantastic}  \\ 
  $4$ & H  & GOD & SS &  FC~\citep{byali2020flash} & MNIST & FLASH~\citep{byali2020flash} \\
  $3/4$ & H  & GOD & SS &  LR~\citep{koti2021swift} & MNIST & SWIFT~\citep{koti2021swift} \\ 
  $4$ & H  & GOD & GC/SS &  LeNet~\citep{koti2021tetrad} & MNIST & Tetrad~\citep{koti2021tetrad} \\    
  \bottomrule
\end{tabular}
\begin{tablenotes}
\item[1] ``\ding{56}'' denotes not support; ``H'' denotes honest majority; ``D'' denotes dishonest majority; ``LR'' denotes logistic regression; ``PR'' denotes Poisson regression; ``FC'' denotes a fully connected neural network. 
\item[2] For the accuracy of the network structure, it is recommended to refer to the original paper. When work is performed on multiple models and datasets, we list only the most complex models and datasets in their experiments. Different works may tailor the network structure or dataset according to the characteristics of the designed FSFT algorithm, such as replacing the loss function, reducing the number of samples, \emph{etc}\onedot
For the accuracy of the experiments, it is recommended to refer to the original paper. 
\end{tablenotes}
\end{threeparttable}
}
% \vspace{-2.0em}
\end{table*}

Under the two server architecture, SecureML~\citep{mohassel2017secureml} first introduces SS technology into model training for machine learning, and designed a secure SGD algorithm for semi-honest models by fusing GC. The effectiveness of SecureML is demonstrated by secure training of linear regression, logistic regression, and fully connected neural network on MNIST.  Subsequently, Quotient~\citep{agrawal2019quotient} is designed by converting the original neural network into a three-valued neural network, \emph{i.e}\onedot, the parameters contain only $\{-1,0,1\}$. The efficiency of SecureML is improved by a factor of 13 while maintaining the model performance. In addition, ABY2.0~\citep{patra2021aby2} reduces the communication overhead of the online phase of the secure multiplication operator by proposing new SS semantics. On top of this, the overall efficiency of SecureML has been improved by up to 6.1 times.  Furthermore, ~\citet{kelkar2022secure} achieves security computation of the exponential operator by transforming SS between different semantics and completes the training of \textit{Poisson regression}.

Under a three-server architecture, ABY3~\citep{Mohassel2018ABY3} designs the FSFT algorithm under both semi-honest and malicious settings by fusing SS and GC. By optimizing the secure multiplication operator and the conversion protocol between SS and GC, the efficiency of SecureML is greatly improved. Experimental results show that ~\citep{Mohassel2018ABY3} is 55,000 times faster than ~\citep{mohassel2017secureml} in secure neural network training. Subsequently, Blaze~\citep{patra2020blaze} optimized ABY3 by proposing new SS semantics in a three-server architecture and achieved an efficiency improvement of up to 2,610 times. Unlike with ABY3 which uses three computational servers, SecureNN~\cite {Wagh2019SecureNN} designs secure FSFT under a semi-honest model by introducing an auxiliary service server. With the assistance of the auxiliary server, SecureNN greatly reduces the communication and computation overhead caused by HE or OT in the preprocessing phase of FSFT under the two-service architecture. Experimental results show that secure neural networks trained using SecureNN over LAN and WAN are 79 and 553 times faster than SecureML, respectively. Subsequently, Falcon~\citep{wagh2020falcon} is designed by fusing the techniques of ~\citep{Wagh2019SecureNN} and ~\citep{Mohassel2018ABY3}. By implementing secure computation of batch normalization functions, ~\citep{wagh2020falcon} can accomplish secure training of complex neural network models including VGG16.
In addition, ARIANN~\citep{ryffel2020ariann} also designs the FSFT algorithm by using the auxiliary server model.
However, by resorting to the function secret sharing technique, ARIANN greatly optimizes the online efficiency of SecureNN nonlinear operations, i.e., achieves secure comparison operations with only one round of communication in the online phase.
Under a four-server architecture,
Privpy~\citep{li2019privpy} improves the efficiency of multiplicative online computation greatly by introducing replicated 2-out-of-4-SS. Furthermore, ~\citep{byali2020flash, chaudhari2019trident, dalskov2021fantastic, koti2021swift} improve the efficiency or security of the FSFT, respectively. 

In summary, the FSFT algorithms achieve complete protection of the training process compared to the PSFT Algorithm. Specifically, all sensitive data in the training process including intermediate parameters are effectively protected by the FSFT algorithms. Meanwhile, by using outsourced computing, they are making full use of the computing resources of the cloud server and reduce the local computing of the participants. Although the efficiency of FSFT algorithms have been greatly improved in recent years, however, they are still very slow and cannot complete most deep learning training tasks in a reasonable time. Therefore, they are suitable for simple training tasks with high security.

% Unlike PSFL algorithms that need to distinguish data distribution, \emph{i.e}\onedot, the dataset is divided by samples or by features, FSFL algorithms can be applied to arbitrarily distributed data and can guarantee the security of both training data and models. 

\begin{figure*}
    \centering
    \includegraphics[width=0.8\textwidth]{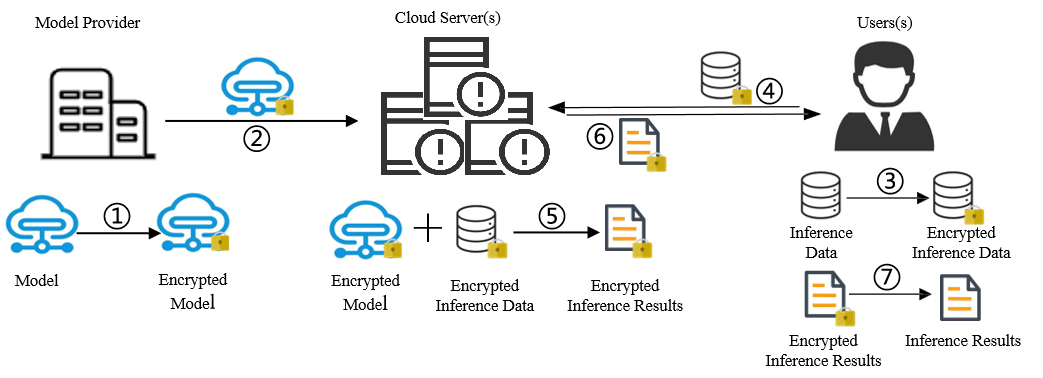}
    \caption{The pipeline of secure federated inference.}
    \label{fig:SFI}
    % \vspace{-1.8em}
\end{figure*}

% 介绍安全联邦预测算法
\vspace{0.1cm}\noindent\textbf{Secure Federated Inference~(SFI).}
The SFI algorithms allow prediction tasks to be completed while protecting the security of the model provider's model and the user's prediction data. Specifically, they convert the operations in the prediction process, such as matrix multiplication and activation functions, into secure operations through secure computing techniques. They often adopt a client-server architecture. In detail, the model provider first deploys the model encrypted or publicly to a cloud server to provide inference services. Users with inference needs send the encrypted inference samples to the cloud servers. Then, the cloud server completes the inference task on the encrypted model and data, and returns the inference results to the user. The pipeline of SFI is shown in~\figref{fig:SFI}. We also classify the SFI algorithm by the number of servers. A summary of the SFI algorithms is shown in~\tabref{table: SFI}. 

Under one server architecture, CryptoDL~\citep{gilad2016cryptonets} and CryptoNets~\citep{hesamifard2017cryptodl} implemented deep convolutional neural networks for secure prediction on MNIST using HE. By using polynomials to approximate nonlinear activation functions, such as ReLU, Sigmoid, \emph{etc}\onedot, they achieve the expected prediction performance of 99\% on MNIST. 
Subsequently, to address the problem that HE is difficult to achieve secure computation of nonlinear operations, ~\citet{fenner2020privacy} proposes a way of server-user interaction. Specifically, the server first completes the linear computation of SFI in the ciphertext state and sends the computation result to the user. The user decrypts and then completes the nonlinear computation in the plaintext state and sends the computed result encrypted to the server. THE-X~\citep{chen2022x}, also adopts the same server-user interactive computation and uses linear neural networks to achieve the approximation of nonlinear operations, such as softmax and layer-norm. After optimization, THE-X achieves the first secure inference of BERT-tiny.
Under two server architecture, DeepSecure~\citep{rouhani2018deepsecure} uses GC to achieve secure prediction and avoids polynomial approximation computation. Although~\citep{rouhani2018deepsecure} uses a binarized neural network to optimize communication efficiency, the communication overhead is still huge. As one of the most famous working SFI,  MiniONN~\citep{liu2017oblivious} is designed by integrating HE and GC. By exploiting the advantages of different techniques, \emph{i.e}\onedot, HE for linear computation and GC for nonlinear computation, MiniONN greatly improves the overall efficiency of the SFI algorithm. Received inspiration from MiniONN Gazelle~\citep{juvekar2018gazelle}, also designes the SFI algorithm by combing HE and GC and additionally improved the computational efficiency of HE by using single instruction, multiple data (SIMD). In addition, Delphi~\citep{mishra2020delphi} improves the efficiency of the SFI algorithm by transferring the computation on HE in Gazelle to be done offline and introducing neural architecture search (NAS) to optimize the preprocessing. It is worth mentioning that Delphi is one of the earlier works that tried to accelerate SFI computation using GPUs. Furthermore, XONN~\citep{riazi2019xonn} reduces the communication overhead by introducing model pruning and secret sharing techniques based on DeepSecure.  Subsequently, Cryptflow2~\citep{rathee2020cryptflow2} optimizes the communication efficiency of the nonlinear activation function of the SFI algorithm using OT and the first ResNet50 and DenseNet121 security predictions were implemented by using Cryptflow2. Unlike existing SFI algorithms, which all use the same bit-width integers for computation, SIRNN~\citep{rathee2021sirnn} uses shorter bit-widths for part of the computation, further reducing the SFI algorithm communication cost, and designing a non-uniform bit-width SFI algorithm in a two-server architecture. Experimental results show that SIRNN can effectively support secure inference for Heads~\citep{saha2020rnnpool}. As the state-of-the-art work in a two-server architecture, Cheetah~\citep{huang2022cheetah} avoids the expensive rotation operation under the SIMD packing technique by cleverly constructing the mapping between plaintext and ciphertext, and improves the computation and communication efficiency of the linear layer of SFI algorithm. Besides, Cheetah also optimizes the nonlinear computation of SFI algorithm by using slient-OT~\citep{yang2020ferret}. 
% After the optimization, Cheetah can complete the security prediction of ResNet-50 on the image dataset under WAN in $2.5$ minutes.

 %~\citep{zhu2022securebinn} further optimization of the linear and nonlinear layers of the SFI algorithm by using secure three-Input and Gate and three-party-OT on top of ~\citep{riazi2019xonn}, and experiments on several real medical datasets, and the effectiveness of the algorithm is demonstrated experimentally on several real medical datasets.
 
Under three server architecture, Chameleon~\citep{riazi2018chameleon} optimizes the communication in the offline phase of the SFI algorithm and the conversion efficiency of GC and SS in the online phase by introducing an assistance server. Furthermore, through consistency detection and fair reconstruction protocols, Astra~\citep{chaudhari2019astra} implements a secure SFI algorithm under the malicious model. Through different forms of secret-sharing techniques Astra optimizes the communication efficiency of the online phase of the secure multiplication operator. \\

Overall, SFI algorithms are well suited for Machine Learning as a Service (MLaaS) scenarios as they protect sensitive data during model deployment and inference.
Currently, they have been able to implement inference tasks for complex deep models in a reasonable time. In the design of algorithms, the technical route of simultaneously using multiple secure computing techniques to enhance the efficiency of SFI algorithms is gradually becoming mainstream. In addition, SFI algorithms are further enhanced effectively through the incorporation of methods such as model quantization and the interaction computation between client and server.

\begin{table*}[t]
  \caption{Secure Federated Inference Algorithms}
  \label{table: SFI}
\resizebox{\textwidth}{!}{
\begin{threeparttable}
\begin{tabular}{*{7}{c}}
  \toprule
    \multirow{2}*{No. of Servers} & \multicolumn{2}{c}{Security Model} & \multirow{2}*{Secure Defense Methods} &  \multirow{2}*{Model} & \multirow{2}*{Dataset} & \multirow{2}*{References}\\
  \cmidrule(lr){2-3}
  & Semi-honest & Malicious \\
  \midrule
  $1$ & D  & \ding{56}  & HE &  Shallow CNN & MNIST  & CryptoDL~\citep{hesamifard2017cryptodl}   \\
  $1$ & D  & \ding{56} & HE &  Shallow CNN & MNIST & CryptoNets~\citep{gilad2016cryptonets}  \\
  $1$ & D  & \ding{56} & HE &  GP & P. falciparum & ~\citet{fenner2020privacy}  \\
  $1$ & D  & \ding{56} & HE &  BERT-tiny & MNLI & THE-X~\citep{chen2022x}  \\
  $2$ & D  & \ding{56} & GC &  Shallow CNN & Smart-sensing  & DeepSecure~\citep{rouhani2018deepsecure}  \\
  $2$ & D  & \ding{56} & GC/OT/SS/HE &  Shallow CNN & CIFAR-10 & MiniONN~\citep{liu2017oblivious}  \\
  $2$ & D  &  \ding{56} & GC/OT/SS/HE &  Shallow CNN & CIFAR-10  & Gazelle~\citep{juvekar2018gazelle}  \\
  $2$ & D  & \ding{56} & GC/OT/SS/HE &  ResNet-32 & CIFAR-100 & Delphi~\citep{mishra2020delphi}  \\
  $2$ & D  & \ding{56} & GC/OT/SS &  Shallow CNN & CIFAR-10 & XONN~\citep{riazi2019xonn}\\
  $2$ & D  & \ding{56} & HE/OT/SS &  DenseNet-121 & ImageNet & Cryptflow2~\citep{rathee2020cryptflow2} \\
  $2$ & D  & \ding{56} & HE/OT/SS &  DenseNet-121 & ImageNet  &  Cheetah~\citep{huang2022cheetah} \\
  %$2$ & D  & - & OT/SS &  Shallow CNN & Malaria & SecureBiNN~\citep{zhu2022securebinn}  \\
  $2$ & D  & \ding{56} & OT/SS &  Heads~\citep{saha2020rnnpool} & SCUT Head~\citep{peng2018detecting} & SIRNN~\citep{rathee2021sirnn}  \\
  $3$ & H  & \ding{56} & GC/OT/SS &  Shallow CNN & MNIST &  Chameleon~\citep{riazi2018chameleon}  \\
  $3$ & H  & Fairness & GC/OT/SS &  SVM & MNIST & Astra~\citep{chaudhari2019astra}  \\
  \bottomrule
\end{tabular}
\begin{tablenotes}
\item[1] ``\ding{56}'' denotes not support; ``H'' means honest-majority; ``D'' means dishonest-majority. 
\item[2] For the accuracy of the network structure, it is recommended to refer to the original paper. When work is performed on multiple models and datasets, we list only the most complex models and datasets in their experiments. Different works may tailor the network structure or dataset according to the characteristics of the designed FSFT algorithm, such as replacing the loss function, reducing the number of samples, \emph{etc}\onedot
For the sake of the accuracy of the experiments, it is recommended to refer to the original paper. 
\end{tablenotes}
\end{threeparttable}
}
% \vspace{-2em}
\end{table*}

\subsubsection{Secure System}
\label{subsec:secure system} 
With the rapid development of SFL, corresponding SFL systems have emerged one after another. They efficiently support secure training or prediction tasks by integrating different SFL algorithms. In this section, we select some representative ones from them to introduce according to the maturity of the system and the amount of users, \etc. A summary of the SFL systems is shown in~\tabref{table: system security}. 

 CrypTFlow~\citep{kumar2020cryptflow} is an open-source SFL system under the Microsoft Research EzPC~\citep{chandran2019ezpc} project. By using ~\citep{kumar2020cryptflow} users can convert Tensorflow and ONNX models into SFL models. The CrypTFlow focuses on SFI tasks, and uses the security operators in SecureNN~\citep{Wagh2019SecureNN}. In addition, CrypTFlow can implement secure SFI algorithms under malicious models, by using TEE, \emph{i.e}\onedot, SGX. CrypTen~\citep{knott2021crypten} is an SFL system built on PyTorch to efficiently support FSFLT and SFI. By integrating with the generic PyTorch API, CrypTen lowers the barrier to use the SFL system, enabling machine learning researchers to easily experiment with machine learning models using secure computing techniques. PySyft~\citep{ziller2021pysyft} is OpenMined's open SFL system that provides secure and private data science in Python. PySyft supports both PSFLT, FSFLT, and SFI. In detail, the private data is decoupled from model training and inference by leveraging secure computing techniques such as SMPC and HE in Pysyft. FATE$\footnote{FATE: https://github.com/FederatedAI/FATE.}$ is an SFL open-source project initiated by Webank’s AI Department. By implementing multiple secure computing protocols, FATE can effectively support PSFLT, FSFLT, and SFI. Furthermore, With the help of highly modular and flexible scheduling system, FATE has good performance and availability. SecretFlow $\footnote{SecretFlow: https://github.com/secretflow/secretflow.}$ is a unified programming framework initiated by Ant for privacy-preserving data intelligence and machine learning. It provides an abstract device layer consisting of ordinary devices and secret devices, whereas cryptographic devices consist of cryptographic algorithms such as SMPC, HE, TEE, and hardware devices. With a device layer containing secret devices and a workflow layer that seamlessly integrates data processing, model training, and hyperparameter tuning, SecretFlow enables efficient SFT and SFI. Rosetta$\footnote{Rosetta: https://github.com/LatticeX-Foundation/Rosetta.}$ is a TensorFlow-based SFL system.
Specifically, by overloading TensorFlow's API, Rosetta allows converting traditional TensorFlow-based algorithm code to SFL algorithm code with minimal changes. The current version of Rosetta integrates multiple SFL algorithms to support SFT and SFI. TF-encrypted$\footnote{TF-encrypted: https://github.com/tf-encrypted/tf-encrypted.}$ is also a TensorFlow-based SFL system. Unlike Rosetta, TF Encrypted makes the system easier to use by leveraging the Keras API. By integrating the relevant SFL algorithms of SMPC and HE, TF Encrypted can effectively support SFT and SFI.  CryptGPU~\citep{tan2021cryptgpu} and Piranha~\citep{watson2022piranha} are two SFL systems that support GPU-accelerated computing and achieve the desired results. CryptGPU implements GPU acceleration of the SFL system by integer decomposition and then uses a submodule of the floating-point Kernel accelerated computational decomposition in Pytorch. Piranha, on the other hand, implements the integer Kernel directly on the GPU to support accelerated computation of the SFL system. PaddleFL$\footnote{PaddleFL:  https://github.com/PaddlePaddle/PaddleFL.}$ is an open-source SFL system based on PaddlePaddle$\footnote{PaddlePaddle:  https://www.paddlepaddle.org.cn.}$. With PaddlePaddle's large-scale distributed training and Kubernetes' elastic scheduling capability for training tasks, PaddleFL can be easily deployed based on full-stack open-source software.

In summary, the current SFL systems are designed to provide an easy-to-use conversion tool for non-cryptography, distributed systems, or high-performance computing professionals. For the sake of the usage habits of machine learning researchers, existing SFL systems are usually developed on mainstream machine learning frameworks such as Tensorflow$\footnote{TensorFlow: https://tensorflow.google.cn.}$ or Pytorch$\footnote{Pytorch:  https://pytorch.org.}$. Specifically, through overloading the APIs of these deep learning frameworks, the SFL systems can convert a machine learning algorithm code to SFL algorithm code with only minor changes.

\begin{table*}[t]
\centering
  \caption{Summary of SFL System}
  \label{table: system security}
  \resizebox{0.6\textwidth}{!}{
\begin{threeparttable}
\begin{tabular}{*{7}{c}}
  \toprule
 Techniques  & PSFT
  & FSFT & SFI & GPU & Frameworks  &  Systems\\
  \midrule
   SMPC & \ding{56} & \ding{56} & \Checkmark & \ding{56} & Tensorflow &CrypTFlow~\citep{kumar2020cryptflow}\\
   SMPC/TEE & \ding{56} & \Checkmark & \Checkmark & \ding{56} & Tensorflow & Rosetta\\
   SMPC/HE/TEE & \ding{56} & \Checkmark & \Checkmark & \ding{56} & Tensorflow & TF-encrypted \\
   SMPC & \ding{56} & \Checkmark & \Checkmark & \ding{56} &  Pytorch  & CrypTen~\citep{knott2021crypten}\\
   SMPC & \ding{56} & \Checkmark & \Checkmark & \Checkmark &  Pytorch  & CryptGPU~\citep{tan2021cryptgpu}\\
   SMPC/HE & \Checkmark & \Checkmark & \Checkmark & \Checkmark & Tensorflow/pytorch & Pysyft~\citep{ziller2021pysyft}\\
    SMPC/HE & \Checkmark & \Checkmark & \Checkmark & \ding{56} &  Tensorflow/pytorch & FATE\\
    SMPC/HE/TEE & \Checkmark & \Checkmark & \Checkmark & \ding{56} & Tensorflow/pytorch   & SecretFlow\\
      SMPC & \ding{56}  & \Checkmark & \Checkmark &\Checkmark  & PaddlePaddle   & PaddleFL\\
     SMPC & \ding{56} & \Checkmark & \Checkmark &\Checkmark  & NA   & Piranha~\citep{watson2022piranha}\\
  \bottomrule
\end{tabular}
  \begin{tablenotes}
\item ``\Checkmark'' denotes supported; ``\ding{56}'' denotes not supported; ``NA'' denotes not adopted. 
\end{tablenotes}
\end{threeparttable}
}
% \vspace{-1.0em}
\end{table*}

\section{Robustness\label{sec:robustness}}

Robust federated learning (RFL) focuses on defending against threats to model performance during the model training process. Compared with SFL techniques that guarantee the correctness of computing results and protect the system from external attackers, RFL considers the threats from internal. More specifically, there are three main threats in RFL, which can not be defended in SFL techniques. Firstly, Non-IID data samples collected by decentralized and inaccessible FL clients could influence the performance of the federated learning model. Furthermore, Byzantine problems may happen to unreliable clients, causing these clients to upload poisoned or failed local models to the server. Moreover, vulnerable clients could be manipulated by human attackers, and then inject backdoors into FL models. Because these threats happen in the data processing or model training procedure in local clients, the above process can not be prevented by SFL techniques. Hence, additional techniques to guarantee the model performance in a federated learning system and prevent internal attacks are required, which can be summarized as robust federated learning. 

In the background, robustness is a vital component in trustworthy machine learning~\citep{varshney2019trustworthy}, as conventional machine learning models are vulnerable to various failures or attacks. When it comes to distributed machine learning systems, it also suffers unreliable clients and lossy communication channels. Federated learning (FL), as a new distributed learning paradigm with privacy-preserving requirements, will face the same even worse vulnerability in different ways. In this section, we focus on the robust techniques in the global aggregation stage on the FL server and discuss the main challenges in RFL that differ from a centralized and distributed data center. We put little attention to the robust system techniques because they have been summarised in the literature~\citep{shafique2020robust, li2021survey, bonawitz2019towards}. We categorize the threats to robustness FL into three classes (Non-IID issues, Byzantine problems, and targeted attacks) and discuss the defenses against them respectively.  
%The category is presented in Figure \ref{fig:robust}, in which the RFL system techniques have been summarised in the literature\citep{shafique2020robust, li2021survey}. The design of RFL system should fully consider the communication staleness and device or network failure\citep{bonawitz2019towards}. 
In detail, we focus on the robustness to Non-IID data issues in \secref{sec:niid}, the robustness to Byzantine problems in \secref{sec:byzantine}, and the robustness to targeted attacks in \secref{sec:target}.

\subsection{Robustness to Non-IID Data} \label{sec:niid}

The vital methods of addressing the Non-IID data challenges are to improve the ability of FL algorithms against model divergence in local training and global aggregation procedures. With the rapid increase of research interest in FL, FL algorithms for Non-IID issues have been proposed in recent years. Furthermore, a review paper~\citep{zhu2021federated} provides its category about FL on Non-IID data from the categories of Non-IID scenario and system view perspectives. In this section, we provide a different view to categorize the FL algorithms towards Non-IID data issues from a technical view. 

\vspace{0.1cm}\noindent\textbf{Non-IID Data Issue}. Due to the privacy regulations that forbid the FL server from directly manipulating the local privacy data samples, the data samples are collected alone in various devices/institutions where the source distributions can be Non-IID in many ways~\citep{hsieh2020non}. Hence, Non-IID data issues are a fundamental challenge in RFL, which results in model performance degradation compared with distributed learning with IID data. These issues commonly exist in FL applications, which further motivate the research against model performance degradation with Non-IID data. 

Concretely, Non-IID data issues influence the model training performance of FL by causing the divergence of local updates~\citep{zhao2018federated}. Although the authors claim that FedAvg~\citep{mcmahan2017communication} could solve the Non-IID data issue to some extent, studies~\citep{zhao2018federated, hsieh2020non, kairouz2021advances} have indicated that the performance drop of FL in Non-IID settings is inevitable. After each round of federated synchronization, local models share the same parameters but converge to different local optimums due to the heterogeneity in local data distributions. Consequently, the divergence of uploaded local update directions causes worse global aggregation results. This divergence usually continues to accumulate during the FL, slowing down the model convergence and weakening model performance. In detail, the scenarios of Non-IID clients can be categorized into several parties~\citep{kairouz2021advances}. 
\begin{itemize}
    \item \textbf{Feature Distribution Skew}. The marginal distribution of label $\mathcal{P}(x)$ varies across clients. In a handwriting recognition task, the same words written by different users may differ in stroke width, slant, \, etc
    \item \textbf{Label Distribution Skew}. The marginal distribution of label $\mathcal{P}(y)$ may vary across clients. For instance, clients' data are tied to personal preferences - customers only buy certain items on an online shopping website.
    \item \textbf{Feature Concept Skew}. The condition distributions $\mathcal{P}(x|y)$ vary across clients, and $\mathcal{P}(y)$ is the same. For instance, the images of items could vary widely at different times and with different illumination. Hence, the data samples with the same label could look very different.
    \item \textbf{Label Concept Skew}. The condition distributions $\mathcal{P}(y|x)$ vary across clients and $\mathcal{P}(x)$ is the same. Similar feature representations from different clients could have different labels because of personal preferences, and the same sentences may reflect different sentiments in language text. 
\end{itemize}

In real-world settings, the data distribution of clients can be the hybrid case of the above scenarios. Furthermore, the number of data samples may be unbalanced among clients as well. Overall, the Non-IID data issue is a basic threat to model performance. Hence, the RFL approaches to guarantee the performance of the FL model is important.

\vspace{0.1cm}
\noindent\textbf{RFL against Non-IID Data}. Training a robust shared global model for an FL system is the most important target to achieve in RFL. For example, FedAvg aims to train a shared model on decentralized privacy datasets, where the shared model shall generalize all distributions of all datasets. However, it could be hard with decentralized Non-IID data. To overcome the Non-IID data, the basic idea of FL for a shared model is to reduce the inconsistency among clients with Non-IID data. Moreover, depending on the problem formulation and targets, we broadly categorize them into optimization-based and knowledge-based methods. In this section, we focus on the robustness of the shared global model in the literature, where most approaches could be implemented on personalized FL at the same time.

We note that personalized FL \citep{tan2022towards} is another solution for Non-IID data issues, which trains personalized models for heterogeneous clients. Personalized FL is robust to Non-IID data issues in most cases, because personalized models usually are derived from the global model via fine-tuning~\citep{li2021ditto, t2020personalized, huang2021personalized} or knowledge transferring~\citep{bistritz2020distributed, lin2020ensemble, he2020group}. Hence, the personalized model naturally fits the local distributions and inherits the robustness of the global model~\citep{li2021ditto}. As the robustness of the global model also affects the inherited personalized models, we focus on the techniques for training a robust global model in the section.

%In this section, we focus on the robustness of the shared global model in the literature, where most approaches could be implemented on personalized FL at the same time.
% Overall, the basic idea of FL for a shared model is to reduce the inconsistency among clients with Non-IID data. Depending on the problem formulation and targets, we broadly categorize them into optimization-based and knowledge-based methods.

\vspace{0.1cm}
\noindent\textbf{Optimization-based Methods}. The basic idea of optimization-based methods is to reduce the divergence of local updates via distributed convex/non-convex optimization techniques. Typically, the proposed optimization-based methods are based on the optimization framework of FedAvg, and further implement various frameworks from different motivations. In detail, FedProx~\citep{li2020federated} adds a proximal term on the client update procedure to restrict their updates to be close. SCAFFOLD~\citep{karimireddy2020scaffold} uses control variates (variance reduction) to correct for the ``client drift'' in its local updates. FedNova~\citep{wang2020tackling} implements a normalized averaging method that eliminates objective inconsistency while preserving fast error convergence. \citet{reddi2020adaptive} propose a generalized FedAvg named FedOPT including client optimization and server optimization procedure. Then, FedOPT derives FedADAGRAD, FedYOGI, and FedADAM by specializing in global optimizer on the server side. FedDyn~\citep{DBLP:conf/iclr/AcarZNMWS21} proposes a solution for a fundamental dilemma, in that the minima of the local-device level empirical loss are inconsistent with those of the global empirical loss. Furthermore, \citet{reisizadeh2020robust} propose FLRA, a general Byzantine-robust federated optimization approach against distribution shifts in samples located in different clients. In summary, optimization-based methods usually are flexible with promising convergence analysis.

\vspace{0.1cm}
\noindent\textbf{Knowledge-based Methods}. Differing from the optimization formulations, knowledge-based approaches for Non-IID issues are motivated by knowledge-transferring techniques. The main drawback of such approaches is the requirement of a proxy dataset on the FL server. For instance, \citet{jeong2018communication} introduces knowledge distillation and proposed federated distillation (FD). FD exchanges model outputs as opposed to FL based on exchanging model parameters. FedKD~\citep{seo2020federated} presents several advanced FD applications harnessing wireless channel characteristics and/or exploiting proxy datasets, thereby achieving even higher accuracy than FL. \citet{li2019fedmd} demonstrate the model heterogeneity problems in FL, where the models differ in clients. Then, federated model distillation (FedMD), based on transfer learning and knowledge distillation, enables FL for independently designed models. Knowledge distillation usually depends on a proxy dataset, making it impractical unless such a prerequisite is satisﬁed. In the above approaches, the quality of the proxy dataset also affects the performance of the FL model, which make knowledge-based approaches unreliable in real applications. To solve that, FedGEN~\citep{zhu2021data} implements a data-free knowledge distillation approach to address heterogeneous FL, where the server learns a lightweight generator to ensemble user information in a data-free manner, which is then broadcasted to users, regulating local training using the learned knowledge as an inductive bias. 

\vspace{0.1cm}
\noindent\textbf{Clustering-based Methods}. Clustering-based methods address Non-IID data issues via cluster clients with similar data distributions. Clustering-based methods are usually orthogonal with optimization and knowledge-based algorithms. Hence, we can run federated optimization algorithms among the clients in the same cluster, which is also known as clustered federated learning (CFL). CFL is presented in~\citep{sattler2020clustered}, which recursively bi-partition clients into two conflict clients clusters according to the gradients. The clustering-based methods are based on the following assumption~\ref{asp:cfl}. 

\begin{assumption}[Clustered Federated Learning~\citep{sattler2020clustered}] \label{asp:cfl}
There exists a partitioning $\mathcal{C} = \{c_1,\dots,c_K\}, \bigcup_{k=1}^K c_k = \{1,\dots,N\}$ ($N \geq K \geq 2$) of the client population, such that every subset of clients $c_k \in \mathcal{C}$ satisfies the distribution learning assumption (i.e., the data distribution among these clients is similar).
\end{assumption}

The key component of clustering-based methods is the algorithm to privacy-respectively distinguish the data distribution of clients and then conduct clustering on these clients. CFL works~\citep{stallmann2022towards, duan2021flexible, pedrycz2021federated, wang2022federated, xie2020multi} clusters clients using K-means clustering based on client parameters. CFL~\citep{sattler2020clustered, sattler2020byzantine} separates clients into two partitions, which are congruent. FL+HC~\citep{briggs2020federated} uses local updates to produce hierarchical clustering. IFCA~\citep{ghosh2020efficient}/HypCluster~\citep{mansour2020three} implicitly clusters clients by broadcasting different models to them and allowing them to choose which cluster to join based on local empirical loss (hypothesis-based clustering). For the model updating method, CFL~\citep{briggs2020federated, sattler2020clustered, sattler2020byzantine} utilizes FedAvg to train cluster models for each cluster during the cluster model updating procedure, ignoring the fact that knowledge from one cluster may help the learning of other clusters. IFCA  conducts parameters-sharing in feature extractor layers and trains personalized output layers via FedAvg. In addition, FedGSP~\citep{dasfaaZengLYHXNY22} assigns clients to homogeneous clusters to minimize the overall distribution divergence among clusters, and increases the degree of parallelism by reassigning more clusters in each round. This idea was further extended to a hierarchical cloud-edge-end FL framework for 5G empowered industries, namded a FedGS, to improve industrial FL performance on non-IID data~\citep{iotjLiHYKLXN22}.

Overall, current CFL methods focus on how to cluster clients better mostly. They learn federated model within client cluster alone and isolated. However, how to enables the federated learning process more efficient across clusters is another open problem in CFL.

% \begin{figure}
%   \begin{minipage}[t]{0.5\linewidth}
%     \centering
%     \includegraphics[width=\textwidth]{Content/robustness/pictures/benigh.pdf}
%     \caption{Federated learning with benign clients.}
%     \label{fig:side:a}
%   \end{minipage}%
%   \begin{minipage}[t]{0.5\linewidth}
%     \centering
%     \includegraphics[width=\textwidth]{Content/robustness/pictures/malicious.pdf}
%     \caption{Federated learning with unreliable clients.}
%     \label{fig:side:b}
%   \end{minipage}
% \end{figure}

\begin{figure}[t]
\centering
\begin{subfigure}[b]{0.4\textwidth}
         \centering
    \includegraphics[width=\textwidth]{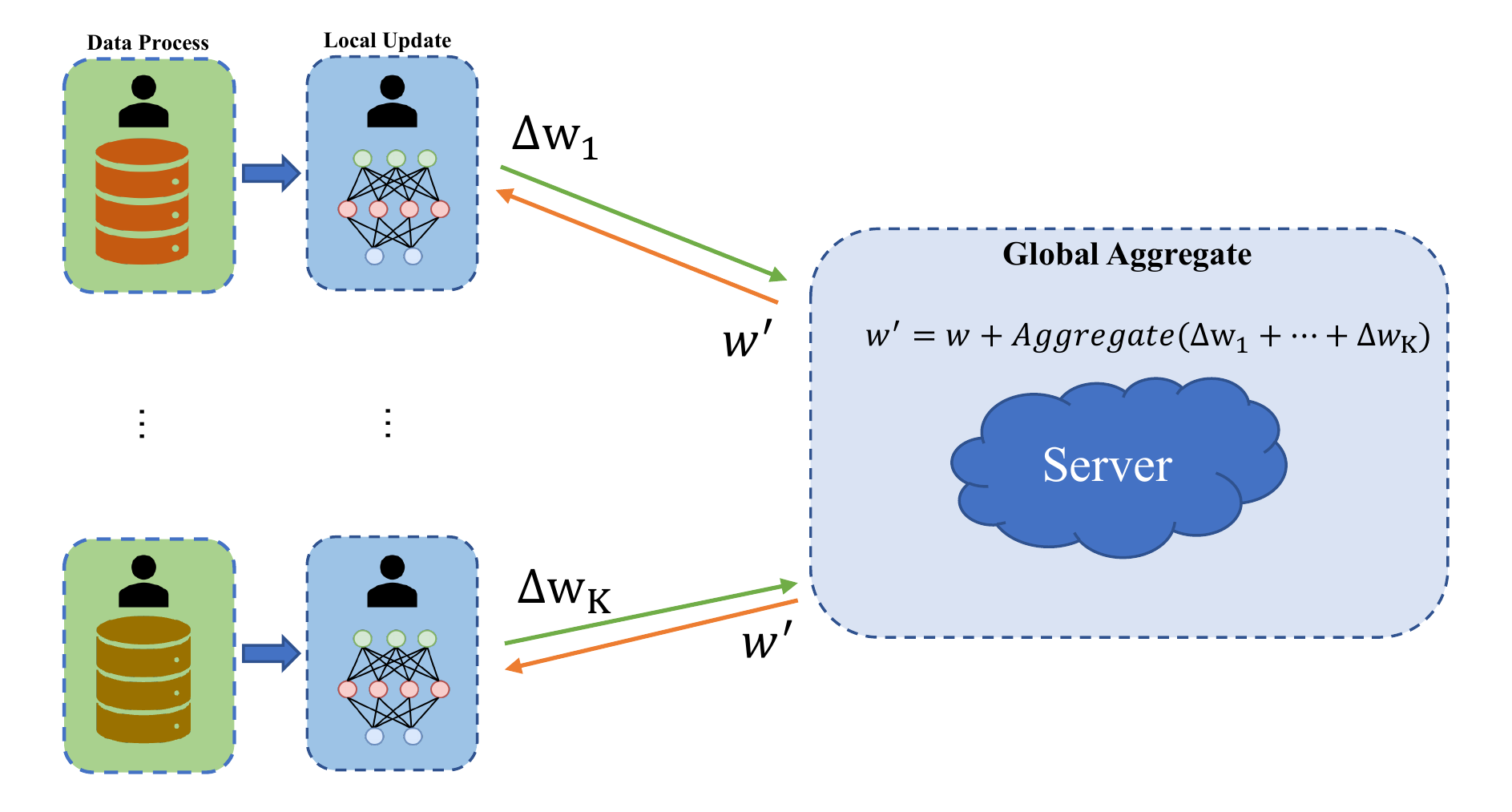}
    \caption{Federated learning with benign clients.}
\end{subfigure}
\hspace{0.5cm}
\begin{subfigure}[b]{0.4\textwidth}
         \centering
    \includegraphics[width=\textwidth]{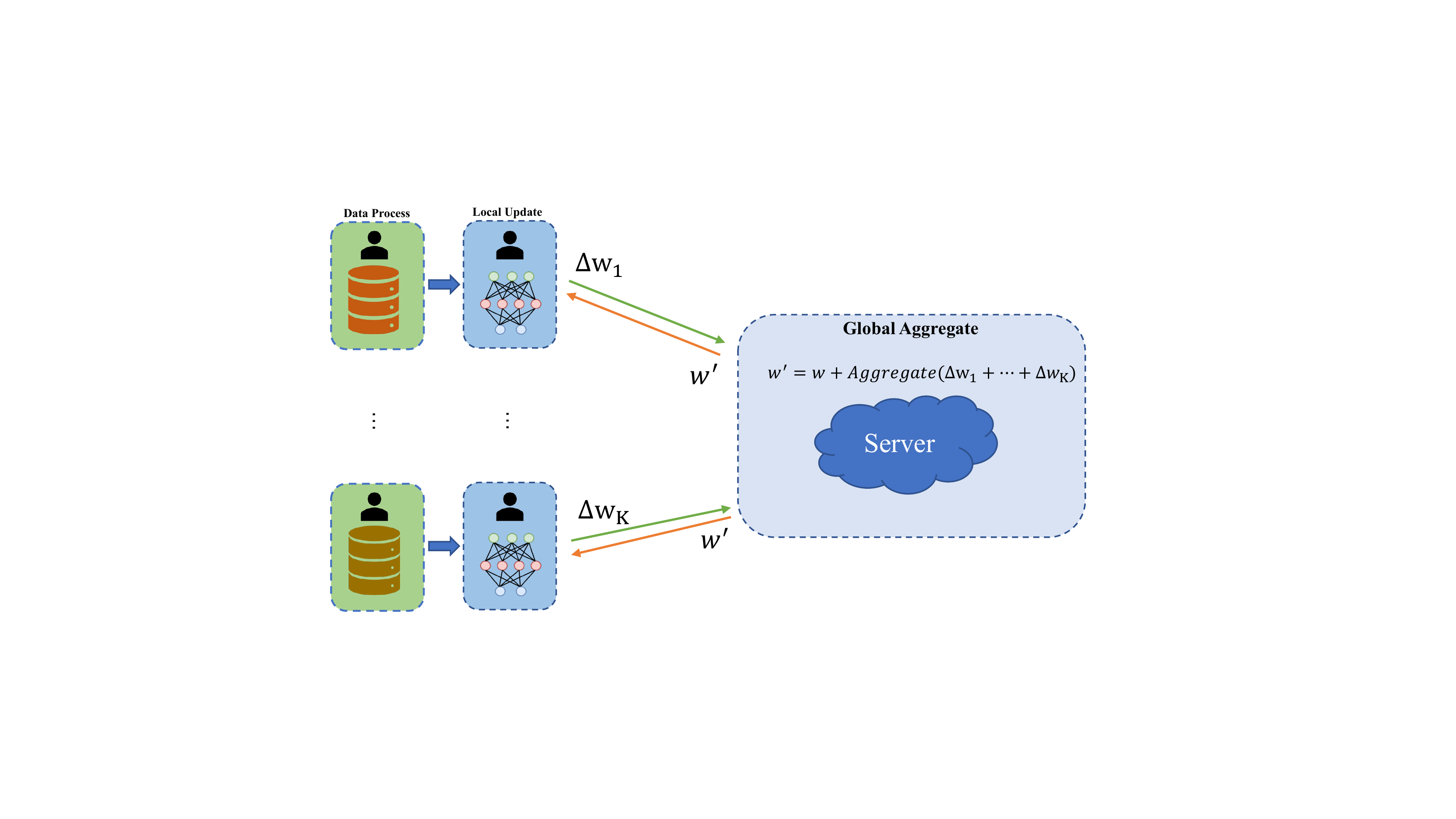}
    \caption{Federated learning with unreliable clients.\label{fig:side:b}}
\end{subfigure}
\caption{The illustration of threats in RFL.}
\vspace{-0.8cm}
\end{figure}

\subsection{Robustness to Byzantine Problem} \label{sec:byzantine}

The failures that happen on the client side may damage the efficiency of the FL system, which can be summarized as the Byzantine problem (the models uploaded from clients are unreliable). In this section, we review the Byzantine problem in FL and summarise the defenses against them. 
FL server is susceptible to malicious clients as a distributed learning paradigm due to the inaccessibility of clients’ local training data and the uninspectable local training processes as shown in \figref{fig:side:b}. In this case, the server cannot be sure whether the clients are honest and benign. The Byzantine problems usually happen during the client update procedure. A subset of clients may be corrupted by an adversary or random failure during the FL. These clients may upload poison local updates to the server. The federated optimization procedure will be ruined if the server unconsciously aggregates these poisoned updates. In this section, an untargeted attack is a specific scenario of the Byzantine problem (clients that may behave abnormally, or even exhibit arbitrary and potentially adversarial behavior). That is, the behavior of a subset of clients is arbitrary as defined in Definition \ref{def:byzantine}. Considering there are $p$ Byzantine clients in an FL system, the affection of the Byzantine attacks on distributed learning can be described as follows:
\begin{equation}
    w = w - \Lambda(\Delta w_1, \dots, \tilde{\Delta w_1}, \dots, \tilde{{\Delta w_p}}, \dots, \Delta w_K), 
\end{equation}
where $\tilde{\Delta w}$ denotes the gradients from Byzantine clients. The Byzantine gradients may cause the optimization direction deviates from the global optimum. Hence, effective Byzantine-robust FL approaches are urgently needed to guarantee the model performance of FL. The FL server could not observe the process that happened on the client side including data processing and client local training. Hence, the information for the server to determine the identity of clients is only the uploaded local updates. Based on that observation, the Byzantine-robust FL that defends the Byzantine problem could be categorized into (1) Robust aggregation, (2) Byzantine detection, and (3) Hybrid mechanism. 

% ~\textcolor{blue}{IK: Should have more definitions!} 
\begin{definition}\label{def:byzantine}
[Byzantine client~\citep{lamport2019byzantine, blanchard2017machine}] A Byzantine client can upload arbitrary local updates to the server
\begin{equation}
\Delta w_i = \begin{cases} *, \;  \quad\quad\quad\quad\quad\,\, \text{if the $i$-th client is Byzantine}, \\ \nabla F_i(w_i; \mathcal{D}_i), \quad \text{otherwise},\end{cases}
\end{equation}
where ``*'' represents arbitrary values and $F_i$ represents client $i$'s objective function.
\end{definition}

\vspace{0.1cm}
\noindent\textbf{Robust Aggregation}. The main goal of robust aggregation approaches is to mitigate the influence of Byzantine clients on global aggregation. Robust aggregation techniques are effective in traditional data center distributed machine learning. A robust aggregation scheme assumes that the poisoned local model updates are geometrically far from benign ones. Hence, methods in the literature aim to build a robust aggregation rule that can mitigate the impacts of malicious attacks to a certain degree. For example in distributed learning, Krum~\citep{blanchard2017machine} and Bulyan~\citep{guerraoui2018hidden} select the local updates with the smallest Euclidean distances to the remaining ones and aggregate them to update the global model. Medoid and Marginal Median~\citep{xie2018generalized} select a subset of clients as a representative set and uses its update to estimate the true center.  GeoMed~\citep{chen2017distributed} estimates the center based on the local updates without distinguishing the malicious from the normal ones. Trimmed Mean and Median~\citep{yin2018byzantine} remove the biggest and smallest values and take the average and median of the remaining ones as the aggregated value for each of the model parameters among all the local model updates. Depending on their aggregation techniques, there are implicit voting majority rules in several methods~\citep{blanchard2017machine, guerraoui2018hidden, yin2018byzantine}, that is, a certain local model update away from others should follow the majority direction. Consequently, they may fail when the number of Byzantine clients is too large. Hence, they have a theoretical break point, indicating the maximum number of Byzantine clients that can be defended as summarised in \tabref{tab:sum}. Overall, robust distributed learning approaches are proposed based on the assumption that the data samples are IID in clients, which conflicts with real-world FL. Besides, the Non-IID scenarios would cause updates to diverge, making the geometric-based robust approaches fail. 

Other robust aggregation techniques mitigate the influence of Byzantine clients via adding regularization terms~\citep{li2019rsa, munoz2019byzantine, xie2020zeno++}. \citet{li2019rsa} enhance the robustness against Byzantine attacks via adding an additional $l_1$-norm regularization on the cost function. Zeno+~\citep{xie2020zeno++} estimates the descent of the loss value in asynchronous SGD and declines the Byzantine clients' contribution via penalty term. \citet{munoz2019byzantine} proposes Adaptive Federated Averaging (AFA) to estimate the quality of the client’s updates and dynamically adjust its averaging weights. Regularization-based methods provide robust analysis under the general cases in optimization and the performance usually is better than geometric-based methods.

In summary, mitigating the influence of poisoned local updates on global optimization procedures is the main idea of robust aggregation approaches. Most of the above robust aggregation schemes suffer from significant performance degradation when a relatively large proportion of clients are compromised, or data among clients is highly Non-IID. Furthermore, recently proposed poisoning attacks~\citep{fang2020local, bhagoji2019analyzing, rong2022poisoning} for FL can bypass these robust aggregation algorithms. Motivated by these new challenges, the Byzantine-robust federated optimization approaches have been proposed~\citep{wu2020federated, pillutla2019robust, portnoy2022towards}, which reveals the further challenges for robust aggregation techniques. From an application view, most of the Byzantine-robust federated optimization approaches assume failure modes without generality. The complex and unpredictable threats~\citep{lyu2020threats} in FL motivate further studies on general Byzantine robustness, which is an open problem.

\vspace{0.1cm}
\noindent\textbf{Byzantine Detection}. Byzantine detection schemes are to identify and exclude malicious local updates so that Byzantine clients will not damage the FL system. Depending on the detection rules of such approaches, we further categorize these methods as validation-based methods and gradient-based methods. For validation-based methods, Error Rate-based Rejection (ERR) rejects local updates that would decrease the global model's accuracy. Loss Function-based Rejection (LFR)~\citep{fang2020local} ranks the model's credibility according to the loss decrease. For gradient-based methods, \citet{li2020learning} use a variational autoencoder (VAE) to capture model-update statistics and distinguish malicious clients from benign ones accordingly. \citet{zhang2022fldetector} observe that the model updates from a client in multiple iterations are inconsistent and propose FLDetector~\citep{zhang2022fldetector} to check their model-updates consistency. Recently, BytoChain~\citep{LiYZLFXS21Byzantine} introduces a Byzantine resistant secure blockchained federated learning framework, which executes heavy verification workflows in parallel and detects byzantine attacks through a byzantine resistant consensus Proof-of-Accuracy (PoA).
In summary, Byzantine detection is compatible with most federated optimization algorithms. Hence, they are more robust than Robust-aggregation schemes. Thus, byzantine detection is a promising direction for exploring robust federated learning. However, these approaches require extra calculation on the server and client~\citep{zhang2022fldetector, zhang2022fldetector, li2020learning}, or additional public data on the server~\citep{fang2020local}. 

\vspace{0.1cm}
\noindent\textbf{Hybrid Mechanism}. Other works combine the above schemes and propose a hybrid defense mechanism. DiverseFl~\citep{prakash2020mitigating} trains a bootstrapping model for each client using some of the client’s local data and compares the trained model with her submitted local model update to examine the local training process. FLTrust~\citep{cao2021fltrust} trains a bootstrapping model on a public dataset and uses cosine similarities between local model updates and the trained bootstrapping model to rank the model’s credibility. CoMT~\citep{han2020robust} couples the process of teaching and learning and thus produces directly a robust prediction model despite the extremely pervasive systematic data corruption. FedInv~\citep{zhao2022fedinv} conducts a privacy-respecting model inversion~\citep{zhu2019deep228} to the local updates and obtains a dummy dataset. Then, FedInv scores and clusters local model updates by Wasserstein distance of the dummy dataset and removes those updates that are far from others. Hybrid mechanism robust federated learning approaches smartly preserve the advantages of both robust aggregation and Byzantine detection techniques, however, usually consume more resources. Hence, the resource-efficient robust approach is also an open problem in applications.

\begin{table*}[t]
\centering
\caption{Summary of representative defenses}
\label{tab:sum}
\resizebox{0.97\textwidth}{!}{
\begin{tabular}{lllllccc}
\hline
\multirow{2}{*}{Category}          & \multirow{2}{*}{Type}               & \multirow{2}{*}{Method} & \multirow{2}{*}{Technique} & \multirow{2}{*}{Break Point} & \multicolumn{3}{c}{Robust to threats}               \\ \cline{6-8} 
                                     &                                       &                         &                            &                              & Non-IID Data & Byzantine Problem & Targeted Attacks \\ \hline
\multirow{8}{*}{Robust Aggregation}  & \multirow{5}{*}{Geometric-based}      & Krum\citep{blanchard2017machine}          & Euclidean distance  &  (K-2)/2K  &  \ding{56}  &  \Checkmark     & \ding{56}       \\
                                     &                                       & BGD~\citep{chen2017distributed}           & Geometric median    &  NA  &  \ding{56}  &  \Checkmark     & \ding{56}          \\
                                     &                                       & \citet{xie2018generalized}               & Geometric median    &  NA  &  \ding{56}  &  \Checkmark     & \ding{56}                  \\
                                     &                                       & \citet{yin2018byzantine}                 & Coordinate-wise median &  K/2  &  \ding{56}  &  \Checkmark     & \ding{56}      \\
                                     &                                       & Bulyan~\citep{guerraoui2018hidden}        & Krum + trimmed median &  (K-3)/4K  &  \ding{56}  &  \Checkmark     & \ding{56}       \\ \cline{2-8} 
                                     & \multirow{3}{*}{Regularization-based} & Zeno++~\citep{xie2020zeno++}              & Inner-product validation  & NA & \ding{56}    & \Checkmark  & \ding{56}     \\
                                     &                                       & AFA\citep{munoz2019byzantine}             & Gradient similarity    &  NA  & \Checkmark    &   \Checkmark     &  \ding{56}  \\
                                     &                                       & RSA\citep{li2019rsa}                      & Loss regularization & NA   &  \Checkmark     &  \Checkmark   &  \ding{56} \\ \hline
\multirow{5}{*}{Byzantine Detection} & \multirow{2}{*}{Validation-based}     & ERR\&LFR~\citep{fang2020local}            & Global validation & NA & \Checkmark & \Checkmark  &  \Checkmark \\
                                     &                                       & Baffle\citep{andreina2021baffle}          & Loss feed-back & NA & \Checkmark   &  \ding{56}   & \Checkmark   \\  \cline{2-8} 
                                     & \multirow{3}{*}{Gradient-based}       & FoolsGold~\citep{fung2020limitations}     & Gradient similarity    &  NA  & \Checkmark    &   \Checkmark      & \Checkmark     \\
                                     &                                       & FLDetector\citep{zhang2022fldetector}     & Gradient consistency   & NA   & \Checkmark     &    \Checkmark   &   \Checkmark   \\
                                     &                                       & \citet{li2020learning}                   & Anomaly detection   & NA   & \Checkmark  &   \Checkmark    &  \Checkmark  \\ \hline
\multirow{5}{*}{Hybrid Mechanism}    & \multirow{4}{*}{\/}       & CoMT~\citep{han2020robust}                & Collaborative teaching & NA &  \Checkmark   &  \Checkmark     &  \ding{56}      \\
                                     &                                       & FLTrust~\citep{cao2021fltrust}            & Global fine-tuning      &  NA   &   \Checkmark   &   \Checkmark      &  \ding{56}   \\
                                     &                                       & FedInv~\citep{zhao2022fedinv}             & Gradient-based clustering &  NA  &  \Checkmark     & \Checkmark       &  \Checkmark     \\
                                     &                                       & DiverseFl~\citep{prakash2020mitigating}   & Filter update  & NA  & \Checkmark   & \Checkmark   &  \Checkmark   \\  
                                     &                                       & \citet{sun2019can}                       & Clipping and DP      & NA & \Checkmark  &  \ding{56}         &  \Checkmark \\ 
                                     &                                       & CRFL~\citep{xie2021crfl}                  & Clipping and smoothing & NA & \Checkmark & \ding{56}  &  \Checkmark \\ \hline
\end{tabular}}
\begin{tablenotes}
\item[1] ``\Checkmark'' denotes supported, and ``\ding{56}'' denotes not supported or not studied.
\end{tablenotes}
% \vspace{-0.5cm}
\end{table*}

\subsection{Robustness to Targeted Attacks}\label{sec:target}

% \begin{table}[h]
% \begin{tabular}{l|c|c|c}
% \hline
% Attacks & Technique & Data poisoning & Model poisoning \\ \hline
% Semantic backdoor~\citep{bagdasaryan2020backdoor}     &           &                &                 \\
% Pixel backdoor~\citep{chen2017targeted}            &           &                &                 \\
% Edge-case backdoor~\citep{wang2020attack}        &           &                &                 \\
% Badnets~\citep{gu2017badnets}                 &           &                &                    \\
% DBA~\citep{xie2019dba}                        &           &                &                 \\ 
% Sybil attack~\citep{fung2020limitations}                                 &           &                &                 \\ 

% \hline
% \end{tabular}
% \caption{Targeted attacks}
% \end{table}

In targeted attacks, the attackers usually could manipulate the learning process of multiple clients and inject specific backdoors into the FL model. In this way. the FL model will output unexpected results on specific inputs with trigger, while the model performance on clean data is normal. Thus, attacks with specific targets are more dangerous threats to RFL. The most discussed targeted attack in the literature is backdoor attacks~\citep{bagdasaryan2020backdoor, gu2017badnets, wang2020attack}, which can be further enhanced by Sybil attacks~\citep{douceur2002sybil, fung2020limitations} in FL. The backdoors in the trained model could be triggered at any time and make the model output unexpected results during the inference stage. The Sybil attacks could manipulate the model training process to control the behaviors of the FL model (backdoor FL models are easier). In this section, we introduce the targeted attacks in FL and review the proposed solutions in the literature.

\vspace{0.1cm}
\noindent\textbf{Backdoor Attacks}. An adversary can conduct complex attacks (\eg, both data poisoning and model poisoning) to implant a backdoor trigger into the learned model. Usually, the model will behave normally on clean data, while predicting a target class if the trigger appears. The backdoor attacks in FL are carried out by adversary clients with smartly organized data/model poisoning attacks. \citet{bagdasaryan2020backdoor} introduce semantic backdoors in FL that cause the model to misclassify even unmodified inputs. Edge-case backdoors~\citep{wang2020attack} force a model to misclassify on seemingly easy inputs that are however unlikely to be part of the training, or test data, \ie, they live on the tail of the input distribution. Badnet~\citep{gu2017badnets} lets Byzantine clients inject label-flipped data samples with specific backdoor triggers into the training datasets. \citet{xie2019dba} propose DBA (distributed backdoor attack), which decomposes a global trigger pattern into separate local patterns and embeds them into the training set of different adversarial parties respectively. \citet{liu2022technical} propose a two-phase backdoor attack, which includes a preliminary phase for a whole population distribution inference attack and generates a well-crafted dataset and the later injected backdoor attacks would benefit from the crafted dataset. If an attacker could inject $c$, fake participants, into the FL system, the FL may suffer from Sybil attack~\citep{douceur2002sybil, fung2020limitations}. A single Sybil attacker could launch arbitrary attacks by influencing the fake participants. For example, Sybil clients contribute updates towards a specific poisoning objective~\citep{fung2020limitations}, which achieves targeted attacks (inject backdoors) and untargeted attacks (ruin model performance). In summary, backdoor attacks are a well-designed combination of poisoning attacks from adversaries. Its form and target could vary in real-world applications, which makes it hard to be detected and defend.

\vspace{0.1cm}
\noindent\textbf{Defense}. The main idea of defending against backdoor attacks is to prevent the formulation of backdoor triggers in model training, as the backdoor attacks can be considered to add an implicit predicting task into the model without noticing and use a trigger to launch a such predicting task in the inference stage. Compared with the defense approaches against the Byzantine problem, targeted attacks are harder to defend as the adversaries may not damage the model performance on clean datasets. Based on the observation that backdoor attackers are likely to produce updates with large norms, several studies are proposed to defend against backdoor attacks by manipulating those suspicious gradients. For example, \citet{sun2019can} suggests using norm thresholding and differential privacy (clipping updates) to defend against backdoor attacks. CRFL~\citep{xie2021crfl} exploits clipping and smoothing on model parameters to control the global model smoothness. A sample-wise robustness certification on backdoors with limited magnitude is proved in the paper. The above techniques clip gradients from all clients to destroy the backdoor tasks. However, the hidden cost is the loss of performance of model training. The targeted attacks are basically conducted via model poisoning and data poisoning. Hence, previously discussed robust aggregation methods and Byzantine detection schemes could defend against backdoor attacks to some extent. Based on that, BaFFLe~\citep{andreina2021baffle} utilizes data information feedback from multiple clients for uncovering model poisoning and detecting backdoor attacks.

\section{Privacy\label{sec:privacy}}

\begin{table*}[t]
\renewcommand{\arraystretch}{1.1}
\caption{A Summary of the Relationship between Privacy Treats and the Existing Methods}
\label{tab:privacy}
\resizebox{\textwidth}{!}{%
\begin{tabular}{llcccccc}
\toprule
\multicolumn{3}{c}{\multirow{3}{*}{\textbf{Privacy Threats}}}                                                 & \multicolumn{5}{c}{\textbf{Defense   Methods}}                                                                                                                                                                   \\ \cmidrule{4-8} 
\multicolumn{3}{c}{}                                                                                 & \multicolumn{2}{c}{Diff. Privacy}                                                    & \multicolumn{2}{c}{Perturbation}                                                               & \multirow{2}{*}{Anonymization}        \\ \cmidrule{4-7} 
\multicolumn{3}{c}{}                                                                                 & \multicolumn{1}{c}{Local Diff. Privacy}  & \multicolumn{1}{c}{Global Diff. Privacy} & \multicolumn{1}{c}{Additive } & \multicolumn{1}{c}{Multiplicative} &         \\ \midrule
\multicolumn{1}{l}{\multirow{2}{*}{\textbf{Data \& Label Leakage}}} & \multicolumn{1}{l}{Parameter Updating} & \citep{li2019end_p_91,yu2019distributed_p_210,zhu2019deep228, zhao2020idlg_p_223} & \multicolumn{1}{c}{\multirow{2}{*}{\citep{bhowmick2018protection_p_10,seif2020wireless_p_155,truex2020ldp_p_174}}}  & \multicolumn{1}{c}{\multirow{2}{*}{\citep{agarwal2018cpsgd_p_3,dubey2020differentially_p_40,naseri2020toward_p_128}}}  & \multicolumn{1}{c}{\multirow{2}{*}{\citep{wei2020framework194,xu2019verifynet_p_198,zhu2019deep228}}}   & \multicolumn{1}{c}{\multirow{2}{*}{\citep{chamikara2021privacy_p_19,gade2018privacy_p_57,jiang2019lightweight_p_81}}}           & \multirow{2}{*}{\citep{song2020analyzing_p_165,xie2019slsgd_p_197}}  \\ \cmidrule{2-3}
\multicolumn{1}{l}{}                                       & \multicolumn{1}{l}{Gradient Updating}        & \citep{mcmahan2017communication121,xu2020information_p_200, hitaj2017deep_s_68, wang2019beyond_s_69, yuan2021beyond_s_70} & \multicolumn{1}{c}{}                     & \multicolumn{1}{c}{}                     & \multicolumn{1}{c}{}                      & \multicolumn{1}{c}{}                              &                      \\ \hline
\multicolumn{1}{l}{\multirow{2}{*}{\textbf{Membership Leakage}}}  & \multicolumn{1}{l}{Training Phase}        & \citep{truex2019demystifying175, mcmahan2017learning123, hitaj2017deep_s_68} & \multicolumn{1}{c}{\multirow{2}{*}{\citep{mcmahan2017learning123, naseri2020toward_p_128, sun2020ldp_p_168}}}  & \multicolumn{1}{c}{\multirow{2}{*}{\citep{hao2019towards_p_70,naseri2020toward_p_128}}}  & \multicolumn{1}{c}{\multirow{2}{*}{\citep{geyer2017differentially_p_63,hao2019towards_p_70,hu2020personalized_p_78}}}   & \multicolumn{1}{c}{\multirow{2}{*}{\citep{chang2019cronus_p_20,feng2020practical_p_52}}}           & \multirow{2}{*}{\citep{choudhury2020syntactic_p_32,seif2020wireless_p_155,su2019securing_p_167}} \\ \cmidrule{2-3}
\multicolumn{1}{l}{}                                       & \multicolumn{1}{l}{Inference Phase}       & \citep{shokri2017membership160, melis2019exploiting} & \multicolumn{1}{c}{}                     & \multicolumn{1}{c}{}                     & \multicolumn{1}{c}{}                      & \multicolumn{1}{c}{}                              &                      \\ \hline
\multicolumn{1}{l}{\multirow{2}{*}{\textbf{Proporties Leakage}}}  & \multicolumn{1}{l}{Individual Proporties}      & \citep{wang2019eavesdrop_s_78, zhang2021leakage_s_13} & \multicolumn{1}{c}{\multirow{2}{*}{\citep{zhao2020local_p_225,zheng2020preserving_p_226}}} & \multicolumn{1}{c}{\multirow{2}{*}{\citep{mcmahan2017learning123,naseri2020toward_p_128,triastcyn2019federated_p_172}}} & \multicolumn{1}{c}{\multirow{2}{*}{\citep{liu2020adaptive_p_110,wei2020federated_p_193,zhu2019deep228}}}  & \multicolumn{1}{c}{\multirow{2}{*}{\citep{feng2020practical_p_52,reisizadeh2020robust}}}          & \multirow{2}{*}{\citep{choudhury2020syntactic_p_32,seif2020wireless_p_155,su2019securing_p_167}} \\ \cmidrule{2-3}
\multicolumn{1}{l}{}                                       & \multicolumn{1}{l}{Population Proporties}     & \citep{mo2020layer,melis2019exploiting_s_80,xu2020subject_s_82,zhu2017unpaired_s_83, chase2021property_p_84} & \multicolumn{1}{c}{}                     & \multicolumn{1}{c}{}                     & \multicolumn{1}{c}{}                      & \multicolumn{1}{c}{}                              &                      \\ \bottomrule
\end{tabular}%
}
\vspace{-0.5cm}
\end{table*}
A privacy attack in federated learning (FL) aims to expose personal information about users who are participating in the machine learning task. This not only endangers the privacy of the data used to train the machine learning models, but also the individuals who voluntarily share their personal information. Initially, it was believed that FL was a distributed machine learning paradigm that could protect personal information. However, the learning process is vulnerable to real-world applications and faces a wide range of attacks. Despite the fact that private data is never shared, exchanged models are prone to remembering the private information of the training dataset. In this section, we present a taxonomy that aims to simplify the understanding of different privacy attacks, as shown in Table~\ref{tab:privacy}.

\subsection{Privacy Threats in Trustworthy Federated Learning}
\subsubsection{Data~\&~Label Leakage}
In the context of Federated Learning, the data and label attack is also commonly referred to as the reconstruction attack. The goal of these attacks is to recover the dataset of a client who participates in an FL task. The attacks typically aim to generate the original training data samples and their corresponding labels. The most common data types that can be targeted by these attacks are images or plain text, which often contain private information owned by the parties.

\vspace{0.1cm}
\noindent\textbf{Gradient-based Data Leakage.}
In a gradient-based Federated Learning (FL) training procedure, as described in~\citep{li2019end_p_91,yu2019distributed_p_210}, selected clients share their gradients with the federated server in communication rounds. However, model gradients may cause a significant amount of privacy leakage as they are derived from the participants' private training datasets. By observing, altering, or listening in on gradient updates during the training process, an adversary (e.g., participants or eavesdroppers) can infer private data using gradients obtained via the training datasets~\citep{zhu2019deep228, zhao2020idlg_p_223}.

To obtain the training inputs and labels, an optimization algorithm designed by \citet{zhu2019deep228} first creates a dummy dataset made up of fictitious samples and labels. Dummy gradients are then obtained using standard forward and backward propagation on the fictitious dataset. The learning process is accomplished by minimizing the difference between the fake gradient and the real gradient. By using this approach, the attacker can infer the training samples and labels with a limited number of training rounds.

\citet{zhao2020idlg_p_223} expanded on the attack proposed in~\citep{zhu2019deep228} by utilizing the relationship between the ground-truth labels and the gradient signs.

\vspace{0.1cm}
\noindent\textbf{Weight-Based Data Leakage.}

In weight-based Federated Learning (FL) frameworks, as described in~\citep{mcmahan2017communication121}, selected clients share their local model parameters with the federated server in communication rounds. Multiple participants have access to the aggregated parameters that the server calculates. A weight update can, therefore, expose the provided data to adversarial participants or eavesdroppers~\citep{xu2020information_p_200}. Malicious participants can calculate the differences between FL models in different update rounds by repeatedly saving the parameters.

\citet{xu2020information_p_200} showed that the model weight can be trained to reveal sensitive information by controlling specific participants during the training phase. In addition, \citet{hitaj2017deep_s_68} offer a Generative Adversarial Network (GAN)-based active attack for recovering training images, where the key to training the GAN is employing the global model as a discriminator. The attacker misleads the target client to release more information about the target label. Furthermore, \citet{wang2019beyond_s_69} have altered the GAN architecture to a multitask GAN.

Apart from GAN-based attacks, \citet{yuan2021beyond_s_70} focus on reconstructing text via model parameters in natural language processing tasks, especially for language modeling.
\subsubsection{Membership Leakage}
Membership inference is a technique used to determine whether a given data sample was part of the training dataset~\citep{li2020label103, lu2020sharing114}. For instance, it can be used to identify whether a particular patient's records were used to train a classifier for a specific disease.

\vspace{0.1cm}
\noindent\textbf{Traing Phase Membership Leakage.}
The work by \citet{shokri2017membership160} focused on the membership inference attack against privacy leakage in machine learning, where an inference model is trained to determine whether a given data sample belongs to the training dataset.

\citet{truex2019demystifying175} extended this approach to a broader context, demonstrating its data-driven nature and high transferability.

Recent studies have identified the gradients and embedding layers as the two areas facing privacy leakage in membership inference attacks~\citep{mcmahan2017learning123}. It has been shown that the embedding of a deep learning network can expose the locations of words in the training data through non-zero gradients, allowing an adversary to conduct a membership inference attack. To address this issue, \citet{hitaj2017deep_s_68} evaluated membership inference attacks against generative models, revealing that many models based on boundary equilibrium GANs or deep GANs are vulnerable to privacy leaks.
 
\vspace{0.1cm}
\noindent\textbf{Inference Phase Membership Leakage}
During the inference phase,~\citet{fredrikson2015model53} developed an inversion method for retrieving private information, revealing that it can expose user-level information.

In a similar vein,~\citet{melis2019exploiting} investigated membership privacy leakage during the inference phase, demonstrating that deep learning models can disclose the placements of words in a batch. In this case, inference attacks are primarily responsible for privacy leaks when attackers can only access model query outputs, such as those returned by a machine learning-as-a-service API~\citep{truex2019demystifying175}.

Furthermore,~\citet{shokri2017membership160} explored privacy leakage during the inference phase by examining the inference membership attack against model query results. In this approach, an inference model is trained to distinguish between training and non-training data samples.

\subsubsection{Property Inference Attack}
This type of attack aims to determine whether an unrelated property of a client or the FL task population is present in the FL model, with the objective of acquiring a characteristic that should not be shared. For example, consider a machine learning model designed to detect faces. An attack might attempt to infer whether training images contain faces with blue eyes, even though this characteristic is unrelated to the primary goal of the model.

\vspace{0.1cm}
\noindent\textbf{Property on Population Distribution.}
This type of attack aims to infer the distribution of a feature in a group of clients. \citet{wang2019eavesdrop_s_78} propose a set of passive attacks using stochastic gradient descent that can be used to infer the label in each training round. To accomplish this, the attacker needs to possess internal knowledge as well as partial external knowledge, which includes information about the number of clients, the average number of labels, and the number of data samples per label.

In a more general FL setting, \citet{zhang2021leakage_s_13} propose a passive attack with limited knowledge to infer the distribution of sensitive features in the training datasets.

\vspace{0.1cm}
\noindent\textbf{Property on Individual Distribution.}
Attackers may attempt to infer the presence of an unrelated property in target clients. With stochastic gradient descent, \citet{mo2020layer} propose a formal evaluation process to measure the leak of such properties in deep neural networks. Both passive and active property inference attacks have been created by \citet{melis2019exploiting_s_80} using internal knowledge, with multitasking learning used to power the active attack~\citep{zhang2021survey}. In a regular FL setting, \citet{xu2020subject_s_82} propose two attacks, passive and active, with the active attack leveraging CycleGAN~\citep{zhu2017unpaired_s_83} to reconstruct gradients using the target attribute. \citet{chase2021property_p_84} suggest a poisoning attack to infer properties, which requires partial internal information to modify the target client's dataset and external knowledge. Finally, in an unusual FL environment using blockchain-assisted HFL, \citet{shen2020exploiting_s_85} propose an active attack.

\subsection{Privacy Defences Method for Trustworthy FL}
\subsubsection{Differential Privacy}

Differential Privacy (DP) allows for information leakage while carefully limiting the harm to people whose private data is stored in a training set. In essence, it hides a person's personal information by introducing random noise. This type of noise is proportional to the largest modification that a single person can make to the output. It should be noted that DP makes the assumption that the adversary has any external knowledge.

\begin{definition}
[Differential Privacy~\citep{rodriguez2023survey}]
A database access mechanism, $\mathcal{M}$, preserves $\epsilon$-DP if for all neighboring databases $x, y$ and each possible output of $\mathcal{M}$, represented by $\mathcal{S}$, it holds that:
\begin{equation}
P[\mathcal{M}(x) \in \mathcal{S}] \leq e^\epsilon P[\mathcal{M}(y) \in \mathcal{S}].
\end{equation}
If, on the other hand, for $0<\delta<1$ it holds that:
\begin{equation}
P[\mathcal{M}(x) \in \mathcal{S}] \leq e^\epsilon P[\mathcal{M}(y) \in \mathcal{S}]+\delta,
\end{equation}
then the mechanism possesses the property of $(\epsilon, \delta)$-DP, also known as approximate DP. In other words, DP specifies a "privacy budget" given by $\epsilon$ and $\delta$. The way in which it is spent is given by the concept of privacy loss. The privacy loss allows us to reinterpret both, $\epsilon$ and $\delta$ in a more intuitive way:
\begin{itemize}
    \item $\epsilon$ limits the quantity of privacy loss permitted, that is, our privacy budget.
    \item $\delta$ is the probability of exceeding the privacy budget given by $\epsilon$ so that we can ensure that with probability $1-\delta$, the privacy loss will not be greater than $\epsilon$.
\end{itemize}
\end{definition}
\noindent and there are mainly two types of DP (\ie, Global Differential Privacy and Local Differential Privacy)

\noindent\textbf{Global Differential Privacy Methods.} 
The global differential privacy scheme has been widely used in many federated learning (FL) methods~\citep{mao2017survey3, choudhury2019differential31, dubey2020differentially_p_40, geyer2017differentially_p_63, hao2019efficient69,  hao2019towards_p_70}. \citet{geyer2017differentially_p_63} presented an FL framework based on global differential privacy by incorporating the Gaussian mechanism to protect client datasets. Specifically, the global model is obtained by randomly selecting different clients at each training round~\citep{geyer2017differentially_p_63}. The server then adds random noise to the aggregated global model, which prevents adversary clients from inferring the private information of other clients from the global model.

However, this framework is vulnerable to malicious servers since they have access to clean model updates from the participants. To address this issue,~\citet{hao2019towards_p_70} proposed adding noise to the local gradients instead of the aggregated model. Following this paradigm,~\citep{geyer2017differentially_p_63} proposed a differential privacy-based privacy-preserving language model at the user level, which achieved comparative performance while preserving privacy. However, one challenge of this method is the trade-off between privacy and utility, as differential privacy inevitably incurs high computation costs.

Overall, the global differential privacy method has an advantage in preserving privacy at a limited cost to model performance. This is because differential privacy is applied to the entire dataset with limited noise, guaranteeing a good statistical distribution.

\vspace{0.1cm}
\noindent\textbf{Local Differential Privacy Methods.}
Various federated learning (FL) approaches have employed the local differential privacy mechanism~\citep{bhowmick2018protection_p_10, cao2020ifed_p_16,liu2020fedsel_p_109, lu2019differentially_p_116, naseri2020toward_p_128,seif2020wireless_p_155,sun2020ldp_p_168,sun2020ldp_p_168,truex2020ldp_p_174,wang2020federated_p_191,wu2021incentivizing_p_195,zhao2020local_p_225, zheng2020preserving_p_226}. The first attempt to combine local differential privacy scheme with deep neural networks was proposed by~\citet{abadi2016deep_p_1}. This privacy-preserving method involves two operations: clipping the norm of updating gradients to limit sensitive information in the data, and injecting noise into clipped gradients. However, this method was not applied to FL systems.

\citep{zheng2020preserving_p_226} investigated local differential privacy and evaluated both efficiency and privacy loss in the setting of FL, ignoring the impact of local differential privacy on model performance. \citet{bhowmick2018protection_p_10} designed a local differential privacy-based method that is free from reconstruction attacks. This approach employs local differential privacy to protect the privacy of samples on the client side and to ensure the privacy of the global model on the server side. In mobile edge computing, \citet{lu2019differentially_p_116} proposed asynchronous FL, which adopts local differential privacy for local model updates. In the Internet-of-Things, \citet{cao2020ifed_p_16} designed an FL framework with local differential privacy as its privacy utility. \citet{truex2020ldp_p_174} scaled the local differential privacy approach to large-scale network training. In the field of natural language processing, \citet{wang2020federated_p_191} used a local differential privacy FL framework for industrial-level text mining tasks and showed that it can guarantee data privacy while maintaining model accuracy.

In summary, due to the nature of local differential privacy methods, they offer a stronger privacy guarantee compared to global differential privacy-based FL methods.

\subsubsection{Perturbation Methods}
They are an alternative approach to provide defenses against privacy attacks that are not based on DP. Its main aim is to introduce noise to the most vulnerable components of
the federated learning, such as shared model parameters or the local dataset of each client, to reduce the amount of information an attacker can extract. There are mainly two types of perturbation methods (\ie, additive based Methods and multiplicative based Methods)

\vspace{0.1cm}
\noindent\textbf{Additive Perturbation Methods.}
Additive perturbation-based FL methods are a popular class of privacy-preserving techniques in FL~\citep{chamikara2021privacy_p_19, feng2020practical_p_52, geyer2017differentially_p_63, hao2019towards_p_70, hu2020personalized_p_78,liu2020adaptive_p_110, triastcyn2019federated_p_172,wei2020federated_p_193, xu2019verifynet_p_198}. These methods aim to incorporate random noise into the weight updates or gradient updates to prevent private information leakage during training. While some methods add noise to the model parameters~\citep{hao2019efficient69, hao2019towards_p_70, triastcyn2019federated_p_172, xu2019verifynet_p_198}, others add noise to the gradient updates~\citep{hao2019efficient69, hao2019towards_p_70, triastcyn2019federated_p_172, xu2019verifynet_p_198}.

One challenge with these methods is to balance the privacy guarantee with the model's accuracy. To address this challenge, \citet{chamikara2021privacy_p_19} propose a lossless data perturbation method that maintains model accuracy while providing a strong privacy guarantee. In this method, random noise is added to the data, rather than the model or gradients.

Other studies, such as \citet{hu2020personalized_p_78} and \citet{liu2020adaptive_p_110}, focus on designing personalized FL models that add noise to intermediate updates or data attributes to ensure privacy. However, one limitation of additive perturbation methods is that they may not be a lossless solution and can affect the model's performance. Additionally, these methods are vulnerable to noise reduction attacks, which can compromise the privacy of the data~\citep{kargupta2003privacy_p_86}.

\vspace{0.1cm}
\noindent\textbf{Multiplicative Perturbation Methods.}
% Instead of adding random noise to data, a multiplicative perturbation that transforms the original data into some space~\citep{chamikara2021privacy_p_19,chang2019cronus_p_20,feng2020practical_p_52, gade2018privacy_p_57, jiang2019lightweight_p_81, zhang2020joint_p_216} is another type of perturbation-based method. Multiplicative perturbation-based FL methods are adapted to the Internet of Things. \citet{chamikara2021privacy_p_19} and \citet{jiang2019lightweight_p_81} use this method to preserve data privacy. In particular, \citet{jiang2019lightweight_p_81} present an FL method for IoT objects. It employs the independent Gaussian random projection to perturb the data of each IoT object. \citet{chamikara2021privacy_p_19} show a multiplicative perturbation mechanism in fog devices. However, since the perturbation is generated by the central server, this method can be vulnerable to insider attackers (\ie, an honest but curious server). \citet{gade2018privacy_p_57} use multiplicative perturbations to obfuscate the stochastic gradient update and protect the gradients from the curious attacker. \citet{chang2019cronus_p_20} and \citet{zhang2020joint_p_216} propose a multiplicative perturbation method for weight update–based FL frameworks. The method is applied to local weight with the aim of preventing the gradient leakage to the servers.
% Overall, compared to the additive perturbation-based FL methods,  perturbation-based FL methods guarantee stronger privacy due to the fact it is harder to reconstruct the original data in this case.
Multiplicative perturbation is an alternative method to adding random noise to data in perturbation-based machine learning. This technique transforms the original data into some space~\citep{chamikara2021privacy_p_19,chang2019cronus_p_20,feng2020practical_p_52, gade2018privacy_p_57, jiang2019lightweight_p_81, zhang2020joint_p_216}, and has been adapted for use in the context of the Internet of Things (IoT) in federated learning (FL) systems. Several studies have explored the use of multiplicative perturbation to preserve data privacy in FL, including \citet{chamikara2021privacy_p_19} and \citet{jiang2019lightweight_p_81}.

In particular, \citet{jiang2019lightweight_p_81} present an FL method for IoT objects that employs independent Gaussian random projection to perturb the data of each object, while \citet{chamikara2021privacy_p_19} propose a multiplicative perturbation mechanism in fog devices. However, it should be noted that the latter method may be vulnerable to insider attackers, i.e. honest but curious servers.

Other studies have used multiplicative perturbations to obfuscate the stochastic gradient update and protect the gradients from curious attackers. For example, \citet{gade2018privacy_p_57} explore the use of this method to protect the gradients in FL systems. Additionally, \citet{chang2019cronus_p_20} and \citet{zhang2020joint_p_216} propose a multiplicative perturbation method for weight update-based FL frameworks, which is applied to local weights with the aim of preventing gradient leakage to servers.

Overall, perturbation-based FL methods that use multiplicative perturbation offer stronger privacy guarantees compared to those using additive perturbation, as it is more difficult to reconstruct the original data in this case.

\subsubsection{Anonymization-based Method}
While perturbation-based methods are often considered to be strong privacy preservation techniques, some works~\citep{ng2021multi131,geyer2017differentially_p_63} argue that they can still degrade data utility. As a result, anonymization-based methods have been presented to address this concern~\citep{song2020analyzing_p_165,choudhury2020syntactic_p_32, xie2019slsgd_p_197,zhao2021anonymous_p_222}.

For instance, \citet{song2020analyzing_p_165} adopt a GAN-based FL framework with anonymization schemes to protect against user-level privacy attacks.

Similarly, \citet{choudhury2020syntactic_p_32} propose an anonymization method for FL that aims to improve both data utility and model performance. This approach involves a global anonymization mapping process for predicting the deployed FL model. Their evaluation shows that this method provides stronger privacy preservation and model performance compared to DP-based FL methods.

In summary, anonymization-based methods provide strong privacy preservation without sacrificing data utility, and can outperform DP-based methods in terms of FL performance. Therefore, they offer a promising alternative to perturbation-based methods for privacy-preserving FL.

% \vspace{-1cm}
\section{Challenges and Future Directions}
\subsection{Secure of Federated Learning\label{sec:SFL summary}}
Although SFL is currently receiving a lot of attention and making impressive progress in both academia and industry, there are still many challenges that need to be addressed. In this section, we summarize the main challenges currently faced by the SFL algorithm. Based on these challenges, some research directions that we consider valuable are proposed. 
\begin{itemize}
% \vspace{-0.1cm}
    \item  Designing more adaptable PSFT algorithms. The current PSFT algorithm is usually customized for "FedAvg". In order to accommodate more complex data distributions in FL, more aggregation algorithms have been developed, such as "FedPox"~\cite{li2020federated}, "SCAFFOLD"~\cite{karimireddy2020scaffold}. How to customize the PSFT algorithm for these new aggregation algorithms is worthy of future research. 
    \item Design of SFL algorithms for adaptive protection. The objects protected by existing SFL algorithms are usually deterministic. Specifically, the PSFT algorithm only protects the security of local model parameters, but exposes the complete global parameters and final model results. The FSFT and SFI algorithms protect the complete data and model security but greatly reduce the overall efficiency of the algorithm. However, the data and model parameters that need to be protected are often different at different stages or in different application scenarios. It makes sense to design an efficient SFL algorithm that can protect the data and models in the FL system accurately according to the requirements. 
    
    \item How to better integrate secure computing techniques into SFL systems? The current SFL algorithm is usually designed to apply safe computation directly. Only the feature of using the same training data for multiple iterations in the machine learning training process is exploited in \citet{kelkar2022secure} to improve the efficiency of secure multiplication. How to combine more features of machine learning algorithms in the training and prediction process to optimize the efficiency of SFL algorithms is a direction worth investigating in the future.

   \item How to perform better engineering optimization of SFL algorithms? The reason for the use of secure computing techniques makes it difficult for the SFL algorithm to be directly compatible with the current mainstream machine learning frameworks. This leads to the fact that existing engineering optimization methods for machine learning cannot be applied to the SFL algorithm. Only a relatively small amount of work has considered engineering optimization of SFL algorithms, such as using GPUs to accelerate FSFT training. How to make more use of engineering optimization to improve the efficiency of SFL algorithms deserves to be investigated in the future.
\end{itemize}
\subsection{Robustness of Federated Learning}
Robust federated learning cares about the trustworthiness of model training performance. By reviewing the robust techniques in robust federated learning, we conclude the challenges and the feasible future directions as follows:
\begin{itemize}
    \item Non-IID Data issues commonly exist in real-world federated learning scenarios, which makes the local gradients geometrically diverse. Consequently, the geometric-based robust distributed learning approaches fail to maintain the robustness of federated learning. Distinguishing the gradients from Byzantine clients or Non-IID clients would be an open problem for future studies on robust aggregation techniques.
    
    \item From the perspective of defense against targeted attacks, the researchers observed that the norm of the poisoned gradients is usually large. Hence, the commonly used approaches are conducting gradient-clipping or gradient-noising. The possibly model training performance degeneration is unavoidable due to the gradient-clipping/noising on benign gradients. Therefore, another open problem in robust federated learning is to keep the model training performance while defending against potential attacks.
    
    \item In real-world applications, the failure, and attack modes are various. Besides, the threat modes are unknown to the server and can be different over time, which further motivates the study of general robust federated learning approaches.
    
    \item  Despite a few FL benchmarks~\cite{caldas2018leaf, terrail2022flamby} and frameworks~\cite{xie2022federatedscope, DBLP:journals/corr/abs-2107-11621, he2020fedml} having been proposed, we lack a standard benchmark to verify the robust ability of proposed FL approaches, which is future work for the FL community.

\end{itemize}

\subsection{Privacy of Trustworthy Federated Learning}
Despite the tremendous growth of privacy-preserving FL in recent years, this research area still poses significant challenges and offers opportunities for developing existing frameworks and creating new methods to enhance both data privacy and utility. Some open research problems and directions are highlighted below:
\begin{itemize}
    \item Privacy-preserving mechanisms in FL have a trade-off between effectiveness and efficiency. Therefore, it is crucial to comprehend usage scenarios under different privacy requirements and study how to optimize the deployment of defense mechanisms.
    \item Data memorization is a major challenge that requires serious attention since neural network-based federated learning may overfit and memorize sensitive information. Anonymizing the data and model can be a potential solution to this issue. Hence, we believe that developing an effective mechanism for anonymizing training datasets is an essential way to ensure privacy preservation in FL.
    \item Developing hybrid approaches for privacy methods in FL by combining various security techniques such as encryption is advantageous. This is because different defense strategies offer significant advantages in different areas, and it is appropriate to leverage their benefits to advance existing frameworks.
\end{itemize}
Overall, TFL is a thriving research field with numerous approaches and applications. This survey aims to summarize existing advances and trends in TFL, with the goal of facilitating and advancing future TFL research and implementation. From a technical perspective, this survey provides a roadmap for building TFL systems. It begins by defining TFL and presents a general picture of the vulnerabilities in the available literature associated with trustworthiness in FL. The survey then reviews recent improvements in TFL regarding security, robustness, and privacy. It explains the threats, summarizes the known defense mechanisms for establishing trustworthy FL in each aspect, and suggests potential future research routes for these elements. We conclude that the study of TFL is a must-have in trustworthy AI due to its importance. There are still several challenges to be addressed and directions to be explored to identify additional threats or vulnerabilities of TFL and appropriate mechanisms to make it a resilient and robust learning paradigm against those threats.

%\section*{acknowledgements}
%This work was partially supported bythe National Key Research and Development Program of China (No. 2018AAA0100204), a key program of fundamental research from Shenzhen Science and Technology Innovation Commission (No. JCYJ20200109113403826), an Open Research Project of Zhejiang Lab (NO.2022RC0AB04), the Major Key Project of PCL (No. PCL2021A06), and Guangdong Provincial Key Laboratory of Novel Security Intelligence Technologies (No. 2022B1212010005).

% To allow for easy dual compilation without having to reenter the
% abstract/keywords data, the \IEEEtitleabstractindextext text will
% not be used in maketitle, but will appear (i.e., to be "transported")
% here as \IEEEdisplaynontitleabstractindextext when the compsoc 
% or transmag modes are not selected <OR> if conference mode is selected 
% - because all conference papers position the abstract like regular
% papers do.
\IEEEdisplaynontitleabstractindextext
% \IEEEdisplaynontitleabstractindextext has no effect when using
% compsoc or transmag under a non-conference mode.

% For peer review papers, you can put extra information on the cover
% page as needed:
% \ifCLASSOPTIONpeerreview
% \begin{center} \bfseries EDICS Category: 3-BBND \end{center}
% \fi
%
% For peerreview papers, this IEEEtran command inserts a page break and
% creates the second title. It will be ignored for other modes.
\IEEEpeerreviewmaketitle

% if have a single appendix:
%\appendix[Proof of the Zonklar Equations]
% or
%\appendix  % for no appendix heading
% do not use \section anymore after \appendix, only \section*
% is possibly needed

% use appendices with more than one appendix
% then use \section to start each appendix
% you must declare a \section before using any
% \subsection or using \label (\appendices by itself
% starts a section numbered zero.)
%

% trigger a \newpage just before the given reference
% number - used to balance the columns on the last page
% adjust value as needed - may need to be readjusted if
% the document is modified later
%\IEEEtriggeratref{8}
% The "triggered" command can be changed if desired:
%\IEEEtriggercmd{\enlargethispage{-5in}}

% references section

% can use a bibliography generated by BibTeX as a .bbl file
% BibTeX documentation can be easily obtained at:
% http://mirror.ctan.org/biblio/bibtex/contrib/doc/
% The IEEEtran BibTeX style support page is at:
% http://www.michaelshell.org/tex/ieeetran/bibtex/
%\bibliographystyle{IEEEtran}
% argument is your BibTeX string definitions and bibliography database(s)
\bibliography{main}

\end{document}